\documentclass[lettersize,journal]{IEEEtran}
\usepackage{amsmath,amsfonts}
\usepackage{algorithmic}
\usepackage{array}
\usepackage[caption=false,font=normalsize,labelfont=sf,textfont=sf]{subfig}
\usepackage{textcomp}
\usepackage{stfloats}
\usepackage{url}
\usepackage{verbatim}
\usepackage{graphicx}
\def\BibTeX{{\rm B\kern-.05em{\sc i\kern-.025em b}\kern-.08em
    T\kern-.1667em\lower.7ex\hbox{E}\kern-.125emX}}

\usepackage{xcolor}
    
\usepackage{balance}

\usepackage{pifont}
\usepackage{tabularx}
\usepackage{booktabs}
\usepackage{makecell}
\newcolumntype{Y}{>{\centering\arraybackslash}X} 

\begin{document}
\title{Towards Robotic Dexterous Hand Intelligence: A Survey}

\markboth{Journal of \LaTeX\ Class Files,~Vol.~18, No.~9, September~2020}%
{Towards Robotic Dexterous Hand Intelligence: A Survey}

\author{
\textbf{Weiguang Zhao}\textsuperscript{1,2*},
\textbf{Tian Liang}\textsuperscript{3*},
\textbf{Xihao Guo}\textsuperscript{3*},\thanks{* Equal Contribution}
\textbf{Rui Zhang}\textsuperscript{2$\dagger$},
\textbf{Irwin King}\textsuperscript{4},
\textbf{Kaizhu Huang}\textsuperscript{3$\dagger$}\thanks{$\dagger$ Corresponding authors}\\
\textsuperscript{1}University of Liverpool \quad
\textsuperscript{2}Xi'an Jiaotong-Liverpool University \\
\textsuperscript{3}Duke Kunshan University  \quad \textsuperscript{4}The Chinese University of Hong Kong
}

\maketitle

\begin{abstract}
Robotic dexterous hands are central to contact-rich manipulation, with rapid progress driven by advances in hardware, sensing, control, simulation, and data generation. However, existing studies are often developed under different assumptions regarding hand embodiments, sensory configurations, task settings, training data, and evaluation protocols, making systematic comparison difficult and obscuring the developmental trajectory of the field. This survey provides a holistic review of dexterous hand research from four complementary aspects. First, we present a hardware-level analysis covering actuation, transmission, perception, and representative hand designs, highlighting the key trade-offs in force capability, compliance, bandwidth, integration, and system complexity. Furthermore, we review control and learning methods for dexterous manipulation from a methodological perspective, grouping representative works by major paradigms and tracing their evolution in chronological order. In addition, we consolidate datasets, modality design, and evaluation practices, which enables methodological progress to be interpreted together with the ways in which it is trained, benchmarked, and assessed. Finally, we discuss the major limitations of current dexterous hand research and summarize the corresponding future directions. By connecting hardware analysis, methodological development, data resources, and evaluation, this survey aims to provide a structured understanding of dexterous hand research and to clarify the most important open challenges for future study.
\end{abstract}


\section{Introduction}

A dexterous hand refers to a mechanical hand capable of controllably manipulating an object within the hand by coordinating multi-finger motions and contact force control, without the aid of external supports~\cite{bicchi2000dexteroushand}. Since the late 20th century, when prosthetic-hand designs began incorporating additional degrees of freedom more aligned with natural human hand motion, dexterous hands have entered a stage of rapid development.

In recent years, learning-based control methods, scalable data generation pipelines, the widespread adoption of simulation platforms, and improvements in highly compliant hardware have been proposed to achieve stable performance, enabling high-degree-of-freedom in-hand manipulation for the first time. The rapid advancement of dexterous hands further stimulates the publication of comprehensive and insightful review studies.

Existing reviews examine dexterous hands from several complementary perspectives. Hardware-oriented surveys analyze hand structures, actuation, transmission, and design trade-offs, comparing rigid, tendon-driven, underactuated, and soft hands in terms of anthropomorphism, force capability, and system complexity~\cite{vertongen2020mechanical,pozzi2023actuated,gu2023soft}. Sensing-oriented surveys establish taxonomies of tactile and multimodal perception for in-hand manipulation, emphasizing perception--action coupling and multifunctional tactile sensing~\cite{yousef2011tactile,li2020review,liang2025structure,zhou2022non}. With the rapid growth of learning-based dexterous manipulation, many surveys further focus on reinforcement learning, imitation learning, and interactive learning paradigms, but they often abstract away the hardware and sensing assumptions under which these methods operate~\cite{song2025Learning,an2025dexterous,huang2025human,welte2025interact}. More holistic surveys adopt broader views across hardware, sensing, and control, while scenario-driven reviews derive design requirements for specific applications such as industrial settings or fragile object handling~\cite{firth2022anthropomorphic,wang2025towards,gao2025empower,zhang2025dexterous}. Nevertheless, most prior reviews still emphasize selected layers or subsystems, with limited effort to jointly connect hardware analysis, methodological evolution, datasets and evaluation, and a dedicated discussion of the field's overarching limitations and future directions.

To this end, this survey revisits dexterous hand research from a holistic perspective. Rather than centering the discussion on any single technical trend, we organize the field around four complementary aspects: a hardware-level analysis of dexterous hands, a methodological review of control and learning approaches, a consolidated discussion of datasets and evaluation practices, and a dedicated examination of major limitations together with future directions. This organization enables us to compare representative works under clearer assumptions and to position different research lines within a common analytical framework. To support this review, our literature collection was primarily conducted through the EI and Web of Science databases. We retrieved articles mentioning relevant concepts in fields such as title, abstract, and subject/topic, and then further screened and organized representative works from this large initial corpus for the subsequent survey and analysis.

Specifically, we first provide a systematic hardware-level analysis of dexterous hands, covering actuation, transmission, perception, and representative hand designs, and highlighting the key trade-offs that shape their development in terms of force capability, compliance, bandwidth, integration, and overall system complexity. Furthermore, we review control and learning methods for dexterous manipulation from a methodological perspective, grouping representative works by major paradigms and tracing the developmental trajectory of the field in chronological order, thereby clarifying how different research directions emerged and evolved over time. To complement this analysis, we further consolidate datasets, modality design, and evaluation practices, so that methodological progress can be interpreted together with the ways in which it is trained, benchmarked, and assessed. Finally, we devote a dedicated discussion to the major limitations of current dexterous hand research and summarize the corresponding future directions of the field, with the aim of clarifying the central open challenges and identifying promising avenues for future study. The contributions of this survey are as follows:

\begin{enumerate}
    \renewcommand{\labelenumi}{\arabic{enumi})}

    \item We review dexterous hand research from a holistic perspective, organizing the field around the interplay among hardware, research methods, datasets, and evaluation.

    \item We provide a systematic hardware-level analysis of dexterous hands, covering actuation, transmission, perception, and representative hand designs, and summarize the key trade-offs in force capability, compliance, bandwidth, integration, and system complexity.

    \item We systematically review control and learning methods for dexterous manipulation from a methodological perspective, grouping representative works by major method paradigms and tracing the developmental trajectory of the field in chronological order. We further consolidate datasets, modality design, and evaluation practices to provide a more complete view of how dexterous manipulation research is developed and assessed.

    \item We provide a dedicated discussion of the major limitations of current dexterous hand research and summarize the corresponding future directions of the field.
\end{enumerate}

The remainder of this survey is organized as follows.  Section~\ref{sec:anatomy} presents the anatomy, focusing on degrees of freedom, actuation, transmission, and sensing. Section~\ref{sec:control} reviews control and learning frameworks for dexterous grasping and in-hand manipulation, together with training and transfer strategies. Section~\ref{sec:dataset_eval} summarizes commonly used datasets and evaluation protocols, and discusses what robustness is measured under different settings. Finally, Section~\ref{sec:challenge_future} highlights key bottlenecks and outlines future directions toward robust dexterous hands and robust real-world deployment.

\section{Anatomy of the Dexterous Hand}
\label{sec:anatomy}

This section provides a system-level anatomy of dexterous hands, focusing on how actuation, transmission, perception, and representative hand designs are organized and coupled in practical systems. We first summarize representative actuation principles in Section~\ref{sec:act}, then review common transmission architectures that map actuator outputs to multi-DoF finger motions in Section~\ref{sec:trans}, followed by perception systems for dexterous feedback in Section~\ref{sec:prec}, and finally provide a comparative summary of existing dexterous hands in Section~\ref{sec:anat}.

\begin{figure*}[h]
    \centering
    \includegraphics[width=0.9775\linewidth]{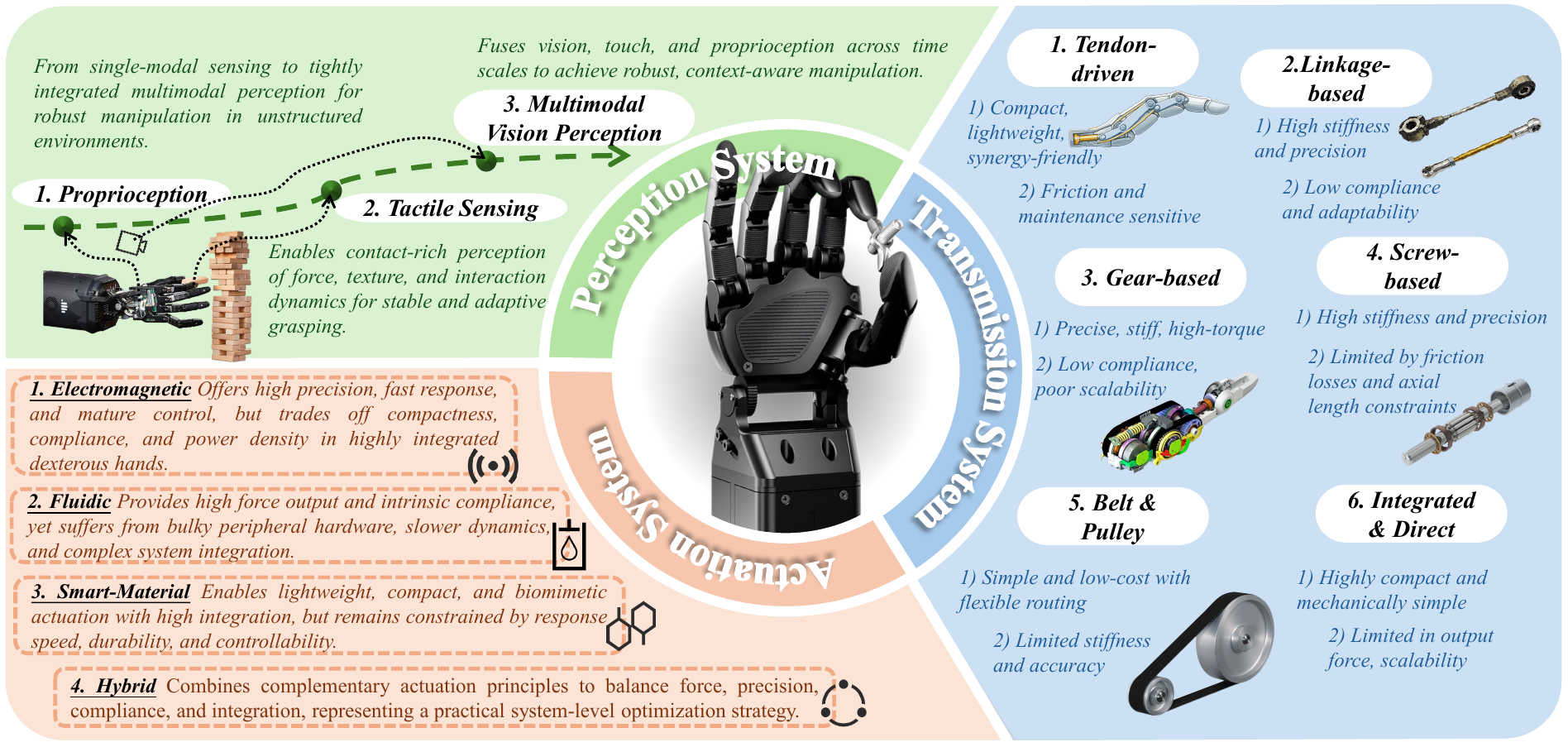}
    \caption{System-Level Architecture of Dexterous Hands: Actuation, Transmission, and Multimodal Perception}
    \label{fig:anotation}
\end{figure*}

\subsection{Actuation System of Dexterous Hands}
\label{sec:act}
An actuation system is the system that converts energy into mechanical motion~\cite{skaar2007def}. Different actuation principles vary fundamentally in their energy sources, force generation mechanisms, achievable power density, dynamic response, and inherent compliance. Consequently, no single actuation solution is universally optimal for all dexterous manipulation tasks. From the perspectives of energy input and physical actuation mechanisms, the actuation systems of dexterous hands can be systematically categorized into four major classes: electromagnetic, fluidic, smart-material, and hybrid actuation.

\textbf{Electromagnetic Actuation (EA).} In electromagnetic actuation for dexterous hands, miniature motors constitute the primary power sources. Coreless DC permanent-magnet motors are widely used due to their high power density, low rotational inertia, and fast dynamic response, enabling precise and rapid finger motion. Frameless torque motors, which eliminate shafts, housings, and bearings, allow direct structural integration into joints, reducing system volume and mass while improving torque transmission efficiency~\cite{zhang2018magneto, guo2024design}. Ultrasonic motors, operating based on the inverse piezoelectric effect, offer advantages such as compact structure, high positioning resolution, and immunity to electromagnetic interference, making them suitable for space-constrained dexterous hands~\cite{guo2023simulation, izuhara2018miniature}. In addition, brushless direct current (BLDC) motors have been adopted in recent large-scale humanoid platforms, such as the Tesla Optimus hand, due to their high efficiency, reliability, and compatibility with high-precision transmission mechanisms.

\textbf{Fluidic Actuation (FA).} Fluidic actuation converts pressurized fluids or gases into mechanical motion and mainly includes hydraulic and pneumatic actuation. Hydraulic actuation relies on pump-powered fluid supply and valve-based flow control to generate motion at the end effector, offering high power density and large output force~\cite{kargov2008development}. To improve compliance, soft hydraulic actuation integrates soft structures with hydraulics, achieving a balance between high force output and flexibility~\cite{zhou2024dexterous}.

Pneumatic actuation, typically realized as pneumatic artificial muscles (PAMs), produces contraction or bending by regulating internal air pressure. Positive-pressure designs include McKibben, fiber-reinforced, and PneuNet actuators~\cite{chou1996measurement, bishop2013force, hadi2023programmable}, while negative-pressure vacuum PAMs (VPAMs) exploit buckling, jamming, or origami-like structures for actuation~\cite{robertson2017new, lee2024design, tawk20193d}. Valves and fluidic circuits serve as essential energy modulation components in fluidic actuation systems, enabling precise, compliant control.

\textbf{Smart-Material Actuation (SMA).} Smart-material actuation exploits controllable material responses under external stimuli to directly generate mechanical output, enabling highly integrated and compact actuation solutions for dexterous hands. Representative examples include shape memory alloys (SMAs), which produce actuation through thermally or electrically induced phase transformations and offer lightweight, compliant, and structurally simple designs, albeit with limitations in response speed and fatigue life~\cite{baek2023dexterous}. Electroactive polymers (EAPs) constitute another important class of smart actuators. Among them, dielectric elastomer actuators (DEAs) generate large deformations and high energy density via Maxwell stress, making them well suited for biomimetic muscle-like actuation~\cite{brochu2010advances, zhang2025review}, while ionic polymer–metal composites (IPMCs) rely on ion migration to achieve low-voltage, highly compliant actuation. More recently, magneto- and photothermally tunable soft magnetic muscles, formed by combining shape memory polymers with magnetic particles, have enabled the integration of actuation and variable stiffness within a single structure~\cite{seong2024multifunctional}. In addition, piezoelectric-ceramic-based robotic hands employing direct joint actuation eliminate the need for complex transmission mechanisms and provide high resolution and fast response, offering a promising pathway for precision dexterous manipulation~\cite{zhang2023piezo}.

\textbf{Hybrid Actuation (HA).} Hybrid actuation refers to systems that intentionally combine multiple actuation principles, or integrate actuation with specific transmission mechanisms, in order to leverage complementary advantages that cannot be achieved by a single actuation modality alone. Typical examples include servo-motor direct actuation coupled with tendon transmission, which combines high control bandwidth with flexible force routing~\cite{zhang2025biomimetic}, and SMA-based actuation integrated with tendons, enabling lightweight and compliant designs while extending actuation range and force delivery~\cite{yang2025lightweight}. Pneumatic–tendon hybrid architectures, such as pneumatic tendon-coupled actuators (PTCA) and discrete pneumatic tendon-coupled actuators (DPTCAs), merge the compliance and variable stiffness of pneumatic actuation with the precision of tendon transmission~\cite{xia2025discrete}. Large-scale humanoid platforms further demonstrate hybrid paradigms, exemplified by the “planetary gearbox + lead screw + tendon” architecture adopted in the Tesla Optimus hand, which balances high force output, precision, and structural compactness. Overall, hybrid actuation systems embody a pragmatic, task-oriented design philosophy that addresses the inherent trade-offs among force, speed, precision, compliance, and integration in dexterous robotic hands.

\subsection{Transmission System of Dexterous Hands}
\label{sec:trans}
Transmission mechanisms define how actuator forces are distributed among joints and contact~\cite{birglen2008underactuated}. While it does not generate energy, the transmission system plays a critical role in determining force efficiency, motion accuracy, compliance, and overall manipulation performance. Transmission mechanisms in dexterous hands can be broadly categorized into tendon-driven, linkage-based, gear-based, screw-based, belt–pulley, and integrated transmission architectures, each offering distinct trade-offs between force capacity, precision, compliance, structural complexity, and suitability for underactuated designs~\cite{palli2014dexmart}.

\textbf{Tendon-Driven Transmission (TDT).} Tendon-driven transmission is a widely adopted approach in dexterous hands, in which flexible tendons are used to transmit force or displacement generated by actuation to the joints, mimicking the muscle–tendon architecture of the human hand~\cite{yin2025disturbance}. Typical implementations combine tendons with pulleys, Bowden cable sheaths, and elastic elements such as springs or rubber components, enabling remote placement of actuators and facilitating underactuated designs. This transmission strategy offers high structural compactness, low mass, and strong suitability for high-degree-of-freedom finger layouts, while naturally supporting motion coupling and synergy. Consequently, tendon-driven mechanisms have been extensively employed in representative systems such as the Shadow Dexterous Hand~\cite{shadowrobot2005}, the Robonaut hand~\cite{bridgwater2012robonaut}, and the DLR David hand~\cite{grebenstein2012hand}. However, their performance is often limited by challenges in tendon tension regulation, friction-induced hysteresis, backlash, and wear, which can degrade control accuracy and response speed and constrain achievable grasping forces. Systematic analyses of tendon-driven and underactuated robotic hands can be found in~\cite{palli2014dexmart, birglen2008underactuated}, while the relationship between the minimum number of tendons and the controllability of multi-DoF hands has been further investigated in~\cite{tu2023posefusion}.

\textbf{Linkage-Based Transmission (LBT).} Linkage-based transmission employs rigid links and revolute joints to directly map actuation outputs into joint motions, thereby establishing deterministic kinematic relationships. Typical implementations include four-bar mechanisms, serial–parallel multi-link configurations, and hybrid screw–linkage structures. Owing to their short force transmission paths and rigid mechanical connections, linkage-based mechanisms can deliver large output forces with high motion repeatability and positioning accuracy, making them well-suited for applications requiring high stiffness and high load capacity. Representative examples include the Schunk SVH five-finger hand~\cite{schunk_svh, ribeiro2023modeling} and the ILDA hand, which achieves multi-DoF finger motion through a sophisticated combination of serial and parallel linkage chains~\cite{kim2021integrated}. However, the inherent rigidity of linkage-based transmission limits compliance and adaptability, while constrained workspaces and complex kinematic designs further limit scalability in high-DoF dexterous hands. Extensive studies have investigated the grasp stability and kinetostatic relationships of linkage-driven hands~\cite{birglen2004kinetostatic, birglen2006grasp}, as well as underactuated linkage designs that enhance adaptability while reducing actuation complexity~\cite{yoon2017underactuated}.

\textbf{Gear-Based Transmission (GT).} Gear-based transmission utilizes gear mechanisms to achieve torque amplification, motion direction change, and coordinated actuation across multiple degrees of freedom. Common implementations include harmonic drives, planetary gear trains, and bevel-gear differential structures. Owing to their high transmission ratios and mechanical rigidity, gear-based transmissions provide accurate positioning and stable torque output, making them suitable for precision manipulation and coordinated joint control. For example, bevel-gear differential mechanisms have been employed to realize multi-DoF metacarpophalangeal (MCP) joint control with enhanced robustness and flexibility~\cite{wan2025rapid}. Earlier dexterous hand designs also extensively adopted gear- and roller-based transmission architectures~\cite{higashimori2005new, quan2013planetary}. However, gear transmissions typically introduce additional mass and volume, which can limit scalability in compact dexterous hands. Moreover, certain high-ratio reducers, such as harmonic drives, are sensitive to impact loads and may suffer from reduced durability under repetitive or high-force manipulation tasks.

\textbf{Lead and Roller-Screw Transmission (LRST).} Lead-screw and roller-screw transmissions convert rotary motion into precise linear displacement and are commonly employed to drive joints or tension tendons in dexterous hands. Owing to their high stiffness and fine positional resolution, screw-based transmissions enable accurate force and motion control and are well suited for closed-loop actuation architectures. Planetary roller screws, in particular, offer improved load capacity and transmission efficiency compared with conventional lead screws, making them attractive for compact, high-force applications. A representative example is the Tesla Optimus hand, which integrates brushless DC motors, planetary roller screws, and tendon-driven mechanisms to achieve high-precision, high-stiffness joint actuation. Screw-based transmission is also adopted in the ILDA hand, where a motor–lead-screw–linkage architecture is used to realize coordinated multi-joint finger motion~\cite{kim2021integrated}. Despite these advantages, screw transmissions suffer from friction-induced efficiency losses and impose constraints on structural length, which may limit their applicability in highly compact or anthropomorphic hand designs.

\textbf{Belt Cable and Pulley Transmission (BCPT).} Belt-, cable-, and pulley-based transmissions utilize flexible belts or ropes routed over pulleys to transmit motion and force from actuators to finger joints. Owing to their simple mechanical structure and low manufacturing cost, such transmission mechanisms were widely adopted in early multi-fingered robotic hands and in designs emphasizing structural simplicity~\cite{tahara2012externally, yuan2020design}. These systems enable flexible routing and decoupling between actuators and joints; however, their relatively low stiffness makes them susceptible to elastic deformation, slip, and backlash, which limits positioning accuracy and force transmission fidelity. As a result, belt- and pulley-based transmissions are less favored in modern dexterous hands that require high precision, high bandwidth, and stable force control, but they remain relevant in low-cost or lightweight implementations.

\textbf{Direct or Integrated Transmission (DIT).} Direct or integrated transmission architectures eliminate or significantly reduce independent transmission mechanisms by tightly integrating transmission functions within the actuation structure itself. In such designs, the deformation of soft chambers or intelligent materials directly generates joint motion, effectively merging actuation and transmission into a single functional unit. Typical examples include soft pneumatic or hydraulic hands, in which the fluidic chambers simultaneously serve as both actuators and transmission pathways~\cite{zhou2024dexterous}, as well as piezoelectric ceramic–driven robotic hands where joints are directly actuated without any intermediate transmission elements~\cite{zhang2023piezo}. These integrated approaches enable compact structures, reduced mechanical complexity, and improved compliance, but often face limitations in output force, scalability, and precise motion control.

\begin{table*}[t]
\centering
\caption{Anatomy of Existing Dexterous Hands}
\label{tab:dex_datasets}

\setlength{\tabcolsep}{2pt}
\renewcommand{\arraystretch}{0.75}

\resizebox{\textwidth}{!}{%
\begin{tabular}{c c c c c c c c c c c c c}
\toprule
\textbf{Hand} & \textbf{Year} & \textbf{Actuation} & \textbf{Transmission} &
\textbf{Tactile} & \textbf{Fingers} & \textbf{DoF} &
\textbf{Voltage} & \textbf{Current} &
\textbf{Capacity} & \textbf{Weight} &
\textbf{Communication} & \textbf{Developer} \\
\midrule

SKKU Hand II & 2006 & EA & GT & \ding{51} & 5 & 10 & 24V & 1.5A & 5kg & 0.9kg & CAN & SKKU \\
DLR-HIT HAND II & 2008 & EA & BCPT & \ding{51} & 5 & 15 & 28V & 3A & 7kg & 1.5kg & PPSeCo, CAN & DLR \& HIT \\
MPL & 2010 & EA & GT & \ding{51} & 5 & 26 & 24V & 6A & 15.9kg & 4.8kg & CAN & JHU \\
DEXHAND & 2012 & EA & LBT & \ding{51} & 4 & 12 & 28V & 3A & 8kg & 4.0kg & EtherCAT, CAN & DLR \\
Self-contained Soft Hand & 2016 & FA & DIT & \ding{51} & 3 & 3 & -- & -- & -- & -- & USB & Harvard \\
Agile Hand & 2018 & HA & GT & \ding{55} & 5 & 16 & 24V & 2A & 5kg & 1.5kg & Proprietary & Agile Robot \\
ShadowHand LITE & 2018 & EA & TDT & \ding{51} & 4 & 13 & 24V & 2.5A & 4kg & 2.4kg & EtherCAT & Shadow Robot \\
ShadowHand Ertra LITE & 2018 & EA & TDT & \ding{51} & 3 & 10 & 24V & 2.5A & 4kg & 2.1kg & EtherCAT & Shadow Robot \\
ShadowHand Super LITE & 2018 & EA & TDT & \ding{51} & 2 & 7 & 24V & 2.5A & 4kg & 1.8kg & EtherCAT, ROS & Shadow Robot \\
HRI Hand & 2020 & EA & TBT & \ding{55} & 5 & 15 & 12V & -- & -- & 0.6kg & Bluetooth & KHU \\
RH56DFX & 2020 & EA & TDT & \ding{55} & 5 & 6 & 24V & 3A & 5kg & 0.5kg & CAN & Inspire Robot \\
Frontiers SMA Hand & 2020 & SMA & DIT & \ding{55} & 5 & 14 & -- & 1.8A & 4.5kg & 0.3kg & Real-time FPGA & Saarland Univ. \\
ILDA Hand & 2021 & EA & LBT & \ding{51} & 5 & 15 & 12V & 0.1A & 9.2kg & 1.1kg & CAN & Ajou Univ. \\
Shadow Hand & 2021 & EA & LBT & \ding{51} & 5 & 24 & 48V & 2.5A & 5kg & 4.3kg & EtherCAT & Shadow Robot \\
UTD SMA Hand & 2021 & SMA & DIT & \ding{51} & 5 & 15 & 6.6V & 0.7A & 0.1kg & 0.2kg & PWM & UT Dallas \\
LEAP Hand II & 2023 & EA & TDT & \ding{55} & 5 & 21 & -- & 0.6A & 2.2kg & 1.0kg & ROS & CMU \\
ARTUS Lite & 2023 & EA & LBT & \ding{55} & 5 & 20 & 24V & 2.5A & 20kg & 1.4kg & CAN & Sarcomere Dynamics \\
DG-5F & 2024 & EA & DIT & \ding{55} & 5 & 20 & 24V & 10A & 20kg & 1.8kg & EtherNet, Modbus & Tesollo \\
Unitree Dex3-1 & 2024 & HA & GT & \ding{51} & 3 & 7 & 58V & 10A & 0.5kg & 0.7kg & USB & Unitree \\
F-TAC Hand & 2024 & HA & BCPT & \ding{51} & 5 & 15 & -- & -- & 2.2kg & -- & CAN & BIGAI \\
GR-2 & 2024 & EA & LRST & \ding{51} & 5 & 12 & -- & -- & 3kg & -- & EtherCAT & Fourier \\
Skilhand & 2024 & EA & BCPT & \ding{55} & 5 & 19 & 24V & 3.5A & 3kg & 0.5kg & EtherCAT & AgiBot \\
XHAND1 & 2024 & EA & GT & \ding{51} & 5 & 12 & 72V & 2.5A & 25kg & 1.1kg & USB & Robot Era \\
DexCo Hand & 2024 & FA & DIT & \ding{51} & 3 & 4 & -- & -- & -- & -- & ROS & UCB \\
DG-4F & 2025 & EA & GT & \ding{55} & 4 & 18 & 24V & 10A & 10kg & 1.4kg & EtherNet, Modbus & Tesollo \\
DG-3F-M & 2025 & EA & GT & \ding{55} & 3 & 12 & 24V & 10A & 5kg & 1.1kg & EtherNet, Modbus & Tesollo \\
Unitree Dex5-1P & 2025 & EA & DIT & \ding{55} & 5 & 20 & 60V & 3A & 4.5kg & 1.0kg & USB & Unitree \\
Optimus Gen 3 & 2026 & HA & LRST & \ding{51} & 5 & 22 & 52V & -- & 8kg & -- & EtherNet & Tesla \\
EFORT Dexterous I & 2026 & EA & GT & \ding{55} & 5 & 15 & 24V & 5A & 10kg & -- & EtherNet,ROS & EFORT \\

\bottomrule
\end{tabular}%
}
\end{table*}

\subsection{Perception System}
\label{sec:prec}
The perception system of a dexterous hand is the system that collects data from onboard sensors. Related algorithms then process this information to perceive both the hand itself and its external environment. It determines the adaptability and intelligence of the dexterous hand in unstructured environments. The perception system has evolved from single-sensor to multiple-sensor use, and then to a “proprioception–tactile–vision” tri-cooperative fusion. It is now rapidly advancing toward multimodal fusion, with technological trends characterized by high precision, miniaturization, multimodality, anti-interference, and low cost.

\textbf{Proprioception.} Proprioception in dexterous hands refers to the capability of sensing the internal state of the hand itself, including joint position, motion, force/torque, and tendon tension, thereby providing essential feedback for closed-loop control, motion accuracy, and protective behaviors~\cite{tuthill2018proprioception}. Compared with external proprioceptive approaches based on vision~\cite{morgan2021towards} or motion capture systems~\cite{shi2017dynamic}, embedded proprioceptive sensing offers higher precision, lower latency, and stronger robustness against occlusion and environmental disturbances~\cite{zhao2023divide,zhao2025bfanet}, and thus constitutes a fundamental component of dexterous hand perception systems.

From a functional perspective, embedded proprioception in dexterous hands can be categorized into four types. Joint pose sensing, which measures joint angles and configurations to support accurate kinematic modeling and position control, is commonly implemented using rotary encoders, potentiometers, Hall-effect sensors, or optical encoders~\cite{li2019development}. Furthermore, motion-state sensing focuses on capturing dynamic information such as velocity and acceleration. Inertial Measurement Units (IMUs) are widely used for this purpose, enabling real-time estimation of hand motion and dynamic states, often in combination with encoder-based velocity sensing~\cite{saegusa2010self}.

In addition, joint force and torque sensing provides direct feedback on interaction forces and internal loads, which is critical for compliant manipulation and safety; such sensing is often realized using strain-gauge-based torque sensors or multi-axis force/torque sensors~\cite{liu2008multisensory}.  Moreover, tendon tension sensing, which is particularly important for tendon-driven and underactuated dexterous hands, enables indirect estimation of joint torques and contact forces by monitoring tendon forces in real time~\cite{friedl2011fas}. Fiber-optic sensing technologies, such as Fiber Bragg Grating (FBG) sensors, have been increasingly adopted due to their high precision and strong immunity to electromagnetic interference.

Together, these proprioceptive sensing modalities provide complementary internal feedback that supports precise motion control, stable force regulation, and robust interaction with uncertain environments. By forming a reliable internal feedback loop, proprioception serves as the foundation for dexterous manipulation performance and plays a crucial role in enabling adaptive, safe, and intelligent hand behaviors.

\textbf{Tactile Sensing.} Human skin contains approximately 17,000 mechanoreceptors, enabling rich tactile perception of contact forces, texture, temperature, and spatial distribution. Inspired by this biological capability, the tactile perception system in dexterous hands provides detailed sensing of object–hand interactions, playing a critical role in stable grasping, manipulation, and fine force regulation~\cite{fang2022tactonet,liu2024tactclnet}. Modern robotic tactile sensing technologies can be broadly classified into resistive, capacitive, piezoelectric, triboelectric, optoelectronic, fluidic, and vision-based approaches, each exhibiting distinct trade-offs in sensitivity, bandwidth, spatial resolution, robustness, and integration complexity. As systematically reviewed in~\cite{wan2017recent}, no single tactile sensing modality can simultaneously optimize all performance metrics, motivating the continued development of multimodal tactile sensors and application-specific designs for dexterous manipulation tasks.

\textbf{Multimodal Vision Perception.} Multimodal perception fusion is a fundamental enabler for dexterous hands to approach human-level manipulation intelligence, driving the evolution from single-modal sensing toward tightly integrated perception of proprioception, touch, and vision. By jointly exploiting complementary sensory modalities, multimodal perception provides more comprehensive and accurate awareness of both the hand’s internal state and external environment, thereby supporting precise, robust manipulation in unstructured scenarios.

At the sensor level, multimodal sensors represent an important trend, offering higher integration density and faster response by simultaneously measuring multiple physical quantities. Recent advances in materials and fabrication have enabled sensors capable of jointly sensing strain, curvature, contact force, temperature, texture, or proximity, such as stretchable optical waveguide sensors~\cite{zhao2016optoelectronically}, multifunctional flexible tactile sensors~\cite{mao2024multimodal}, textile-based soft sensors~\cite{seong2025soft}, and fabric-based multimodal capacitive sensors~\cite{song2025fabric}. These designs reduce sensing redundancy while enhancing spatial and temporal coherence across modalities.

At the algorithmic level, multimodal fusion can be performed at the data, feature, and decision levels, or through hybrid strategies combining all three. Representative systems integrate tactile and proprioceptive sensing to emulate human mechanoreceptors and enable fine texture recognition~\cite{rostamian2022texture}, or fuse vision, touch, and joint sensing to achieve reliable multimodal perception despite heterogeneous sensing latencies, as demonstrated by the RAPID Hand~\cite{wan2025rapid}. Hybrid fusion strategies further balance control performance and computational cost by hierarchically combining low-level motion–force features with higher-level visual object representations.

Beyond perception alone, multimodal sensing is increasingly co-designed with actuation and control to form closed-loop perception–decision–control architectures. Examples include indirect sensing of tendon length and tension via encoders or vision~\cite{ogahara2003wire, gilday2020vision}, vision-centric multimodal hands that minimize onboard sensors while retaining rich perception~\cite{chen202518}, and monolithic soft electronic skins that integrate multimodal sensing with neuromorphic signal processing and closed-loop actuation~\cite{wang2023neuromorphic}. Together, these developments highlight a clear trend toward tightly coupled, multimodal perception systems that enhance dexterity, adaptability, and system-level efficiency in next-generation robotic hands.

\subsection{Anatomy of Existing Dexterous Hands}
\label{sec:anat}
Table~\ref{tab:dex_datasets} summarizes representative dexterous hand platforms and provides a compact anatomy across key hardware and system attributes, including the release year, actuation and transmission choices, developer, communication interface, finger count, DoF, electrical envelope, and basic capability indicators such as fingertip force, payload capacity, and weight. In this table, the actuation abbreviations follow the taxonomy in Sec.~II-A, where electromagnetic actuation is denoted as EA and is discussed as a mainstream solution due to its mature control and fast response. Similarly, the transmission abbreviations follow Sec.~II-B, where tendon-driven, linkage-based, gear-based, and integrated transmission represent different trade-offs among force efficiency, motion accuracy, compliance, and integration complexity.

Moreover, Table~\ref{tab:dex_datasets} highlights that dexterous hands span a wide range of design envelopes rather than converging on a single best architecture. This variation directly shapes what is feasible in closed-loop manipulation, because actuation and transmission determine the achievable torque and bandwidth, while the communication interface and electrical constraints influence controller update rate, sensing integration, and system latency. In addition, the large spread in fingertip force, payload capacity, and weight provides an intuitive reference for discussing robustness-oriented design.

\section{Research on Dexterous Hand}
\label{sec:control}

\begin{figure*}[ht]
    \centering
    \includegraphics[width=0.9775\linewidth]{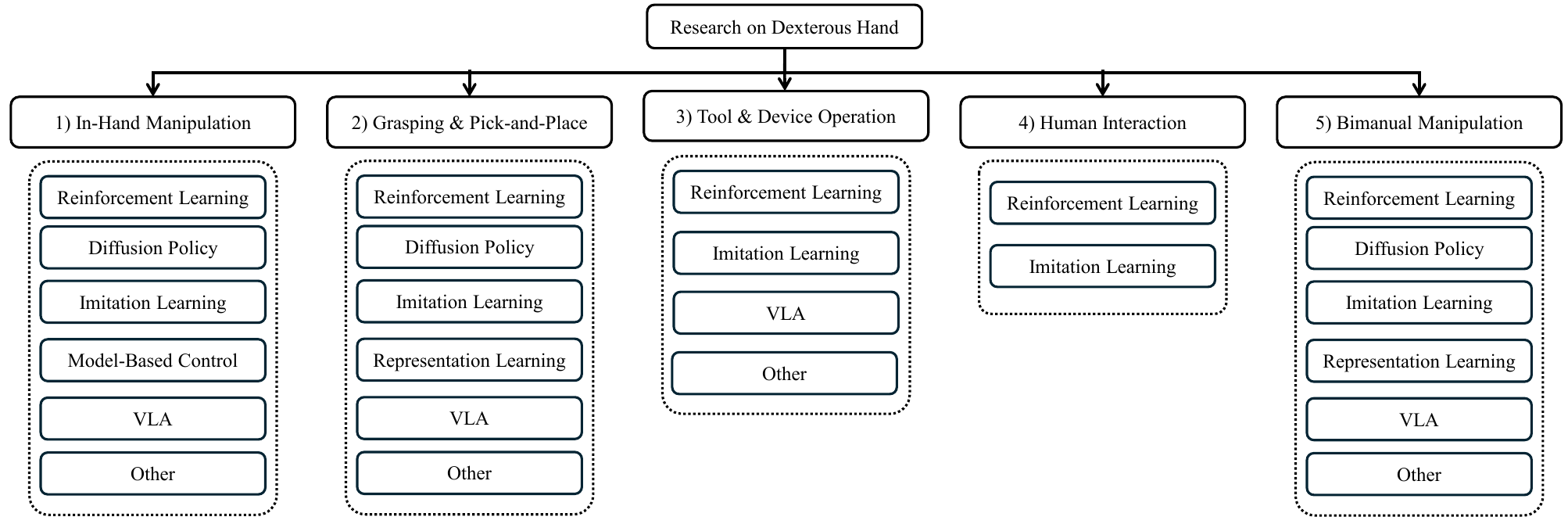}
    \caption{Research on Dexterous Hand}
    \label{fig:research}
\end{figure*}

Dexterous hand manipulation comprises heterogeneous tasks with diverse contact dynamics, force requirements, and perceptual demands, making universal training paradigms impractical. Consequently, task-oriented analysis is necessary to clarify how different training paradigms support dexterous manipulation across varying task regimes. In this section, we first summarize a general open-ended workflow shared by dexterous-hand tasks in Section~\ref{sec:rese}, and then examine the current progress of dexterous-hand research from the perspective of specific task categories, as shown in Fig.~\ref{fig:research}, including In-Hand Manipulation (Section~\ref{sec:innhand}), Grasping \& Pick-and-Place (Section~\ref{sec:grasp}), Human Interaction (Section~\ref{sec:human}), Tool \& Device Operation (Section~\ref{sec:tool}), and Bimanual Manipulation (Section~\ref{sec:bim}).

\subsection{Research Workflow}
\label{sec:rese}
Dexterous hand control aims to reproduce the fine manipulation skills of the human hand, enabling stable grasping, precise motion control, and safe interaction with objects. The central challenge is to translate the high degrees of freedom, compliant contacts, and rich multi-modal perception of the human hand into robotic control strategies that satisfy real-time constraints, stability, and safety requirements. With recent advances in bionics, material science, and artificial intelligence, dexterous hand systems have gradually evolved from monolithic controllers to hierarchical and learning-augmented frameworks. In this section, we summarize a three-stage closed-loop research workflow for dexterous hand control and introduce its key mechanisms and algorithmic components.

\begin{figure}
    \centering
    \includegraphics[width=1\linewidth]{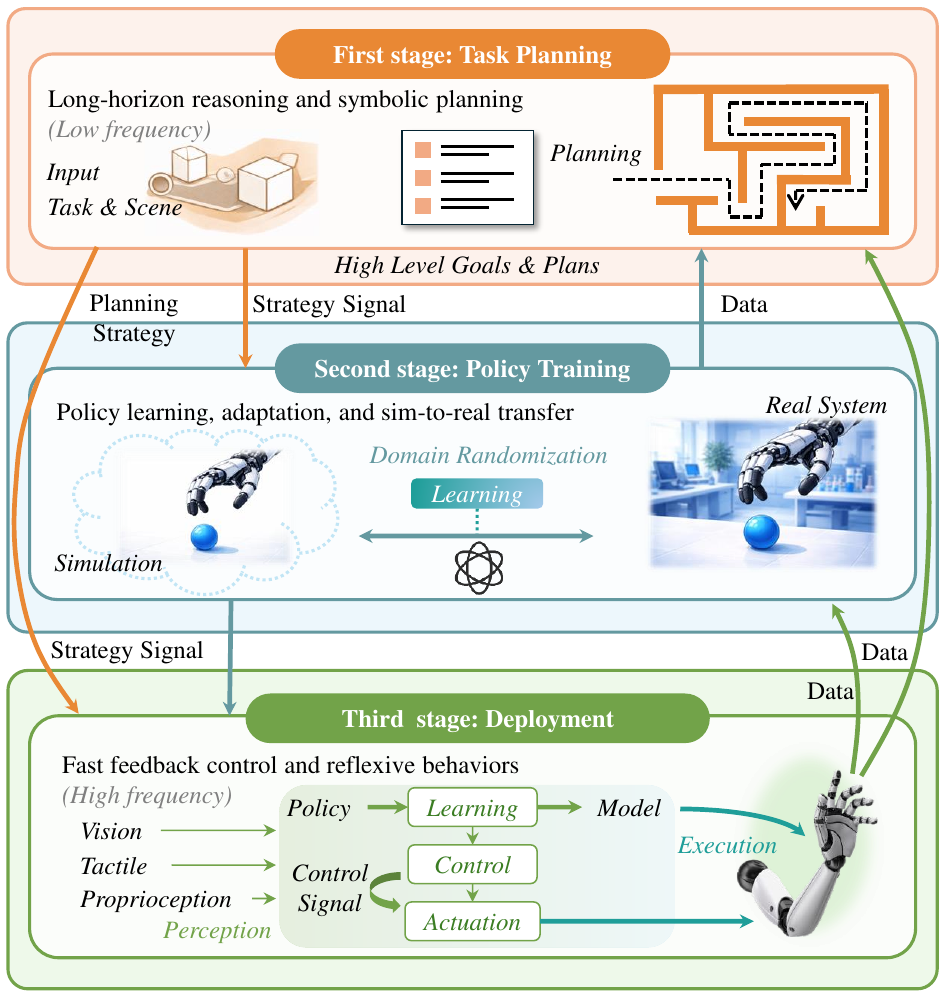}
    \caption{Research Workflow on Dexterous Hand}
    \label{fig:workflow}
\end{figure}

Dexterous hand control systems are commonly formulated as closed-loop pipelines that integrate perception, planning, execution, and evaluation. To cope with the high dimensionality, hybrid contact dynamics, and task diversity of dexterous manipulation, hierarchical structures are widely adopted, as they improve data efficiency, modularity, and generalization across tasks~\cite{huang2019continuous}. Existing designs often decompose control according to abstraction level and temporal scale. At the system level, high-level task planning is decoupled from low-level execution, so task generalization can be separated from policy realization~\cite{bai2025towards}. In learning-augmented systems, high-level planners may further leverage large language models to decompose complex manipulation goals into ordered sub-goals, while low-level controllers map these sub-goals into executable motor commands under current states and constraints~\cite{song2023llm}. Representative hierarchical systems include the DLR-HIT Hand architecture, which integrates low-level control, distributed FPGA-based data processing, high-level control, and external command interfaces~\cite{liu2009modular}, and three-layer designs that separate task planning, event-level coordination, and low-level trajectory and force control~\cite{yu2022dexterous}.

As learning-based control and foundation-model-driven methods become increasingly prevalent, overall system performance depends not only on the control hierarchy itself but also on the research workflow that connects planning, training, and deployment. A widely adopted paradigm can be summarized as a three-stage closed loop, as illustrated in Fig.~\ref{fig:workflow}. In the first stage (task planning), problem discovery and task specification are conducted, including requirement definition, failure analysis, and the formulation of objectives, constraints, and evaluation criteria. In the second stage (policy training), manipulation policies are trained and validated using simulation and collected data, while transfer robustness is improved through domain randomization, calibration, and sim-to-real adaptation. In the third stage (deployment), the learned policy is executed on a real dexterous hand with real-time perception and safety monitoring, and execution logs and failure cases are collected as feedback to refine both the training data and planning, forming an iterative closed loop.

 \subsection{In-Hand Manipulation}
\label{sec:innhand}
\begin{figure*}[h]
    \centering
    \includegraphics[width=0.9775\linewidth]{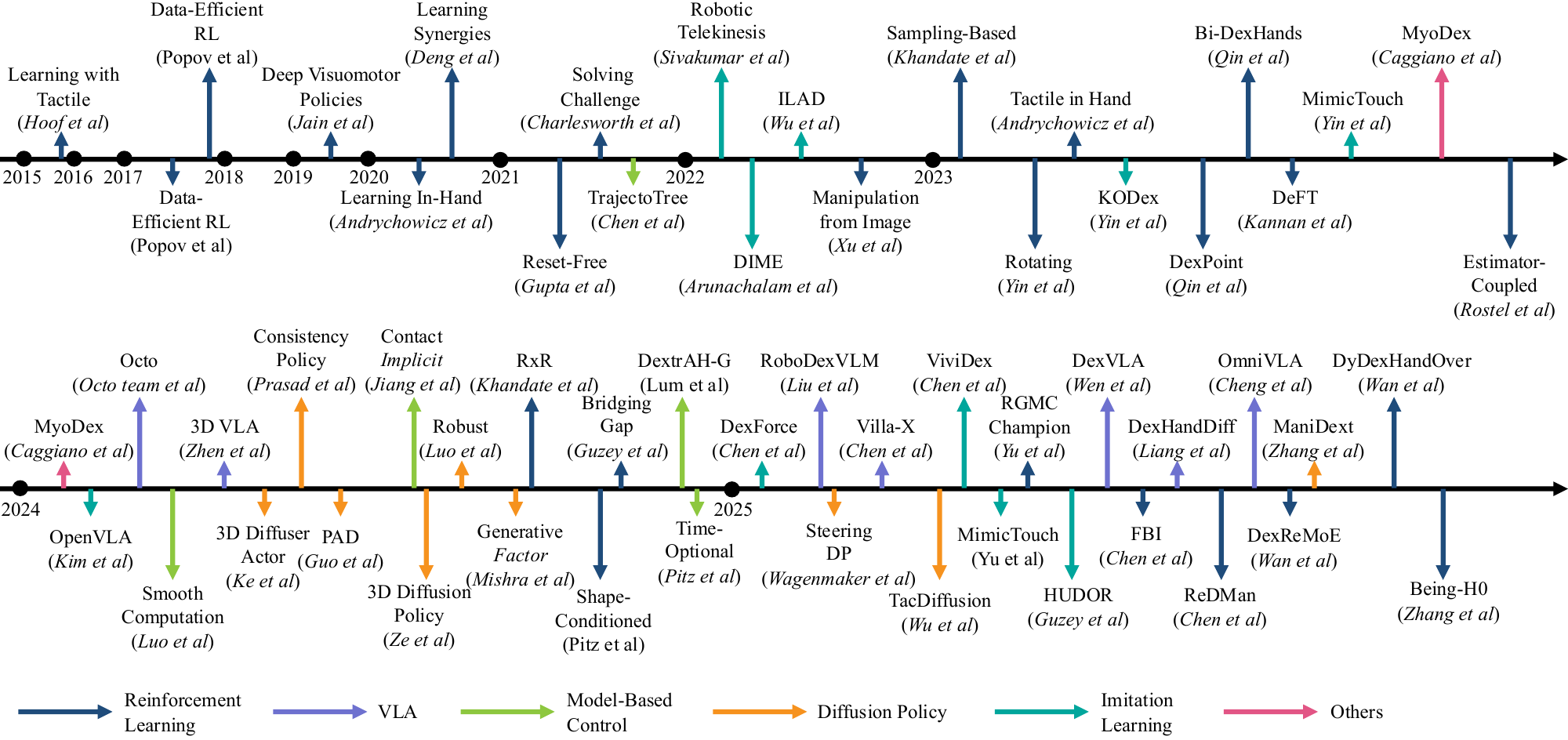}
    \caption{In-Hand Manipulation Timeline}
    \label{fig:In-hand Time Line}
\end{figure*}

In-hand manipulation constitutes the core capability of dexterous hands and serves as the fundamental building block for more complex manipulation tasks. Precise force–motion coupling, severe visual occlusion, and irreversible failure modes make explicit modeling and open-loop planning particularly challenging. Fig.~\ref{fig:In-hand Time Line} shows the research timeline for in-hand manipulation.

\textbf{Reinforcement Learning.} In-hand manipulation involves persistent contact, frequent contact switching, and partial observability, where tightly coupled force–motion dynamics and visual occlusion render explicit modeling and trajectory optimization difficult. Reinforcement learning addresses these challenges by learning closed-loop manipulation strategies directly from interaction, enabling stable in-hand translation, rotation, and reorientation without accurate contact models.

Existing work primarily focuses on improving training efficiency and robustness under high-dimensional contact dynamics. Data-efficient RL methods accelerate convergence~\cite{popov2017data}, while improved sampling strategies address exploration bottlenecks in in-hand motion planning~\cite{khandate2023sampling}. Reset-free learning further enables continuous in-hand manipulation training without manual intervention after failure~\cite{gupta2021reset}. Vision-based RL extends these approaches by decomposing manipulation into guided substeps learned directly from visual observations~\cite{xu2022dexterous}. Point-cloud-based observations improve geometry-aware generalization~\cite{qin2023dexpoint}, while post-simulation fine-tuning enables targeted adaptation to real contact conditions~\cite{kannan2023deft}.

As in-hand manipulation increasingly involves multimodal feedback and coordination, RL frameworks integrate kinematic priors~\cite{ze2023h} and tactile sensing~\cite{yuan2024robot} to stabilize manipulation under partial observability. Extensions to bimanual in-hand manipulation~\cite{chen2023bi} and human–robot handover scenarios~\cite{zhou2025dydexhandover} further demonstrate the scalability of RL in contact-rich, interactive manipulation.

\textbf{Diffusion Policy.} In-hand manipulation demands coordinated multi-finger motion, smooth force–motion transitions, and robustness under contact uncertainty. Diffusion policies naturally fit this setting by generating continuous action trajectories, avoiding abrupt control signals, and enabling diverse, multimodal manipulation behaviors. Recent work explores different ways of integrating diffusion policies into dexterous manipulation. SteeringDP combines diffusion policies with reinforcement learning by optimizing a low-dimensional latent policy whose outputs are decoded into high-dimensional actions, preserving smooth multimodal behaviors while improving task performance~\cite{wagenmaker2025steering}. Several methods adapt diffusion policies to contact and force-sensitive in-hand settings. TacDiffusion generates actions directly in the 6D force–torque space, enabling precise tactile manipulation under geometric and frictional uncertainty~\cite{wu2025tacdiffusion}. DexHandDiff introduces interaction-aware diffusion planning to mitigate ghost states and improve adaptability under changing targets~\cite{liang2025dexhanddiff}. ManiDext conditions diffusion policies on object 3D motion trajectories to synthesize high-quality in-hand manipulation sequences~\cite{zhang2025manidext}. Diffusion policies are also extended to perceptually grounded manipulation. PAD unifies future visual prediction and action generation within a single diffusion process, supporting cross-modal learning from both videos and robot demonstrations~\cite{guo2024prediction}.

\textbf{Imitation Learning.} Imitation learning is well-suited for in-hand manipulation, where high-dimensional finger coordination, persistent contact, and force–motion coupling make exploration and reward design difficult. By leveraging expert demonstrations, IL injects human manipulation priors and enables stable learning of contact-rich behaviors. Early studies focus on low-cost acquisition of in-hand demonstrations from human motion. Robotic Telekinesis retargets monocular human videos to robotic hands without specialized hardware~\cite{sivakumar2022robotic}, while ILAD synthesizes large-scale demonstrations from human grasp affordances to support cross-object generalization~\cite{wu2022ILAD}. 

Subsequent work incorporates structural and dynamical constraints to improve learning efficiency and stability. DIME enables practical RGB-based demonstration collection~\cite{arunachalam2022dexterous}, and KODex models coupled with hand–object dynamics using Koopman operators~\cite{han2023utility}. More recent approaches emphasize contact-aware imitation. DexForce integrates measured contact forces from kinesthetic demonstrations~\cite{chen2025dexforce}, while MimicTouch leverages multimodal tactile demonstrations for force-sensitive manipulation~\cite{yu2025robotic}. Vision-based imitation further evolves toward outcome-centric supervision, where HUDOR extracts object-oriented imitation signals from human videos~\cite{guzey2025bridging}, and ViViDex distills video-derived demonstrations into physically plausible in-hand manipulation policies~\cite{chen2025vividex}.

\textbf{Model-based Control.} Model-based control provides explicit physical reasoning, reliable multi-finger coordination, and strong generalization for contact-rich in-hand manipulation, making it safer, more stable, and less data-dependent than purely learning-based approaches.

TrajectoTree\cite{chen2021trajectotree} combines tree search with contact-implicit trajectory optimization in a hierarchical framework, decomposing and efficiently solving the complex contact-switching problems in multi-contact dexterous manipulation. This leads to faster, more stable, and more physically consistent manipulation planning. SmoothComp~\cite{luo2024robust} proposes a model-based framework that leverages system dynamics for prediction, planning, and robust control, and addresses real-time and stability challenges through tube MPC and a piecewise-linearized optimization structure. LeTac-MPC~\cite{xu2024letac} combines neural tactile encoding with a differentiable MPC layer, forming a learning-augmented model-based control framework that enables stable, interpretable, and real-time tactile-reactive grasping under dynamic disturbances and varying object physical properties. On the other hand, DextrAH-G~\cite{lum2024dextrah} combines a strongly model-driven geometric fabric controller with deep reinforcement learning and policy distillation, achieving safe, coordinated, real-time, and highly generalizable pixels-to-action dexterous grasping.

\textbf{VLA.} VLA models introduce high-level semantic grounding and long-horizon task reasoning into dexterous in-hand manipulation. Rather than directly replacing low-level force and motion control, VLA primarily operates at the high and mid-level, providing task decomposition, semantic constraints, and goal-conditioned guidance that complement contact-rich controllers. OpenVLA~\cite{kim2024openvla} establishes the open-source VLA paradigm by fine-tuning vision–language–action models for robotic manipulation, enabling dexterous hands to execute language-conditioned tasks. Octo~\cite{team2024octo} scales this paradigm to large trajectory datasets and demonstrates cross-embodiment generalization, while subsequent work further expands training data sources by incorporating Internet-scale human interaction data into the VLA framework~\cite{luo2025being}.

Recent studies extend VLA toward stronger geometric grounding and hierarchical control, which are critical for in-hand manipulation. 3D-VLA~\cite{zhen20243d} augments 2D visual inputs with 3D scene representations, enabling reasoning about hand–object geometry and contact. RoboDexVLM~\cite{liu2025robodexvlm} and Villa-X~\cite{chen2025villa} decompose long-horizon instructions into structured subgoals, marking a shift from end-to-end language policies toward hierarchical architectures that interface more naturally with low-level controllers. DexVLA~\cite{wen2025dexvla} introduces plug-in diffusion experts to handle multimodal coordination during reorientation, while OmniVLA~\cite{cheng2025omnivtla} unifies visual, linguistic, and haptic inputs within a single Transformer architecture, improving feedback stability during manipulation.

\textbf{Other.} Several additional works, although not belonging to the categories discussed above, also introduce highly constructive approaches for in-hand manipulation and have significantly advanced the field. MyoDex~\cite{caggiano2023myodex} introduces a generalizable prior for high-DoF musculoskeletal hand models (MyoHand). By jointly training a multi-task policy across a wide range of manipulation behaviors, the method automatically discovers reusable muscle synergies within the complex muscle-driven dynamics. These synergies significantly improve generalization to unseen tasks and novel objects. When used as an initialization for downstream fine-tuning, the MyoDex prior accelerates convergence across 57 dexterous manipulation tasks and consistently achieves higher success rates than single-task training. Robot Synesthesia~\cite{yuan2024robot} presents a visuotactile fusion approach for in-hand manipulation, addressing the limitations of vision- or touch-only methods. Its key idea is the tactile point cloud, which maps tactile signals into 3D point clouds and fuses them with visual point clouds to form a unified touch–vision representation.

\begin{figure*}[ht]
    \centering
    \includegraphics[width=0.9775\linewidth]{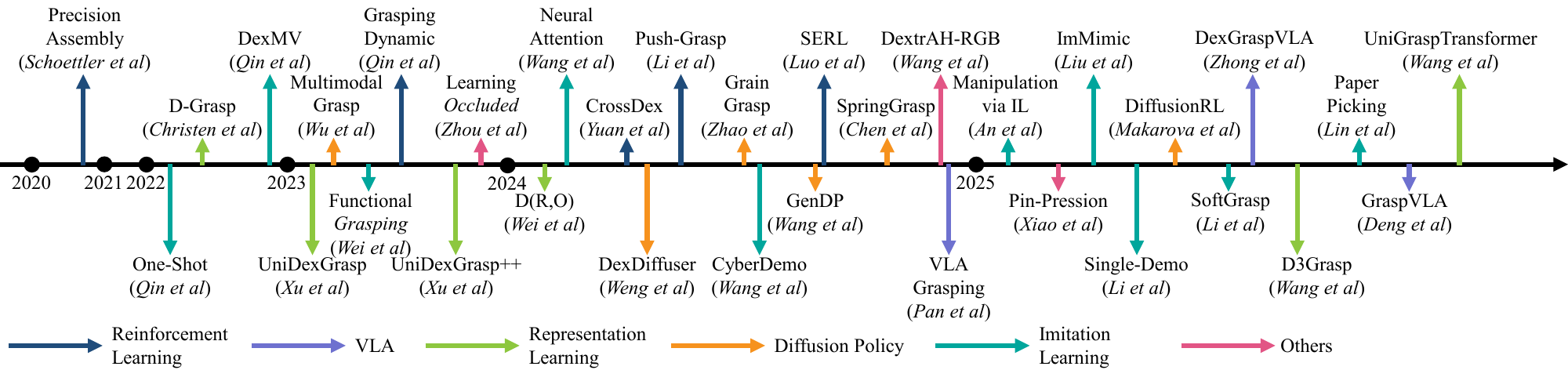}
    \caption{Grasp \& Pick-and-Place Timeline}
    \label{fig:ppt}
\end{figure*}

\subsection{Grasping and Pick-and-Place}
\label{sec:grasp}
Grasping and pick-and-place constitute the foundational skills of dexterous manipulation, establishing stable hand–object coupling for downstream tasks~\cite{bicchi2000robotic,zhao2026tele,li2025language}. Dexterous grasping is typically formulated as grasp generation followed by grasp execution, where feasible contact configurations are selected based on object geometry and task constraints, and then realized through motion planning and feedback control~\cite{huber2024domain, liu2024realdex, song2025overview, li2024comprehensive}. Pick-and-place further introduces perceptual uncertainty and force regulation during transport and placement, requiring adaptive control based on tactile and contact feedback~\cite{billard2019trends}. Fig.~\ref{fig:ppt} shows the research timeline for grasping and pick-and-place.

\textbf{Reinforcement Learning.} Reinforcement learning plays an important role in dexterous hand grasping and pick-and-place tasks, as it enables policies to learn contact-rich manipulation strategies directly through interaction.  DRLKT~\cite{wang2023dexterous} adopt task-related pretraining strategies to accelerate learning, while CrossDex~\cite{yuan2024cross} achieves cross-morphology data transfer between different dexterous hands, eliminating the need for retraining when the platform changes. In addition, Luo et al. focus on improving exploration strategies for real-world dexterous hands, significantly reducing interaction data requirements compared with standard RL~\cite{luo2024serl}.

Generalization across environments and embodiments is another central concern. To improve robustness under visual uncertainty, CrossDex~\cite{yuan2024cross} further extend point-cloud-based task representations to enable generalization across different dexterous hands, providing a promising solution to sim-to-real transfer failures. RL is also applied to increasingly complex grasping and pick-and-place scenarios that demand precise force regulation and rich contact reasoning. Schoettler et al. demonstrate submillimeter-precision industrial assembly using a vision-based RL policy~\cite{schoettler2020deep}, while Li et al. introduce coordinated push–grasp primitives to improve success rates in cluttered environments~\cite{li2024grasp}. Hu et al. employ inverse reinforcement learning to infer manipulation strategies from human demonstrations, enabling robust interaction with dynamic targets~\cite{hu2023grasping}.

\begin{figure}[h]
    \centering
    \includegraphics[width=1\linewidth]{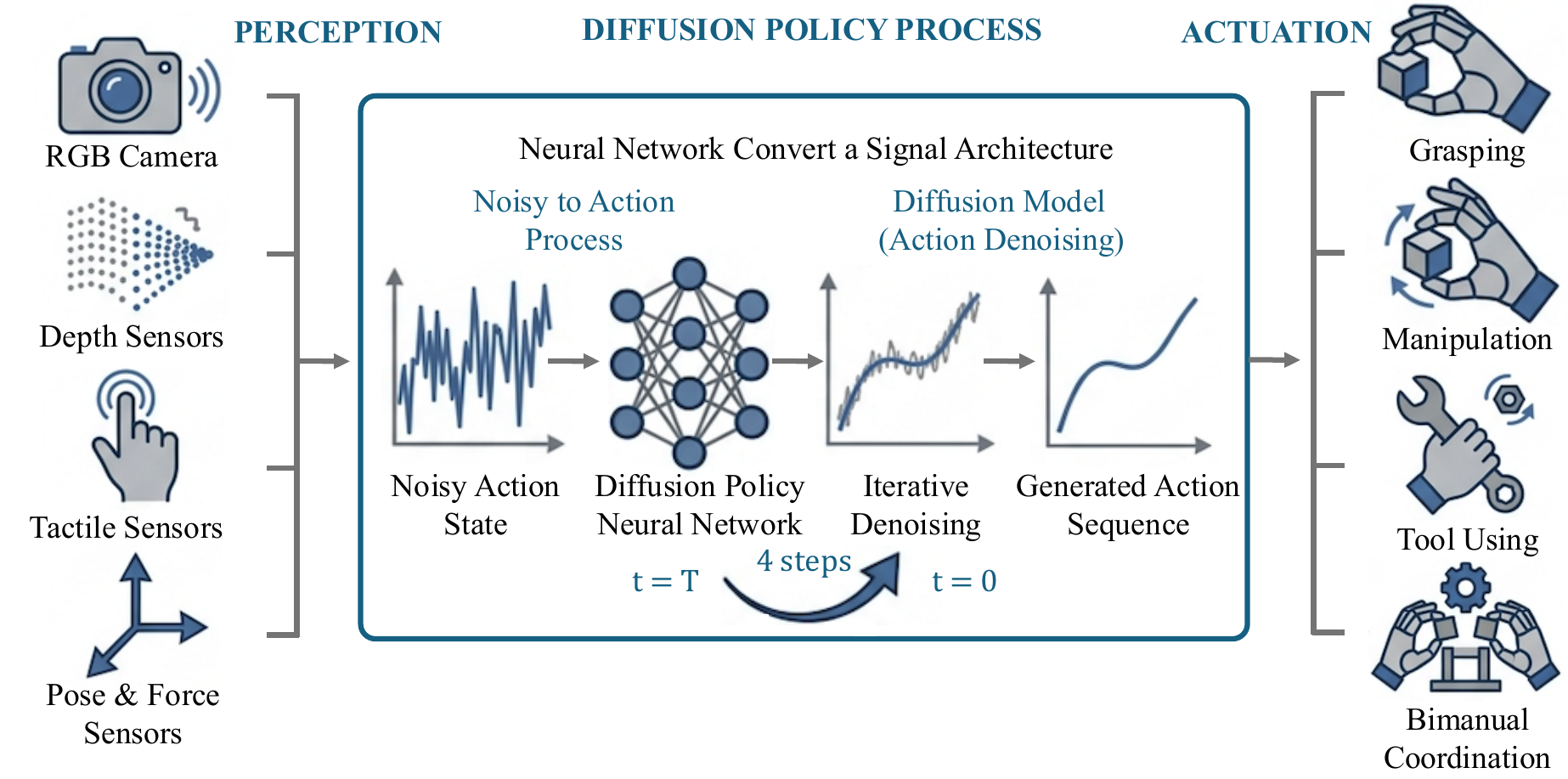}
    \caption{Overview of diffusion policies connecting multimodal perception to diverse dexterous manipulation tasks. The generative policy denoises action states into smooth control trajectories for varying task requirements. }
    \label{fig:dp}
\end{figure}

\textbf{Diffusion Policy.} As shown in Fig.~\ref{fig:dp}, diffusion policies generate smooth, multimodal, and robust action trajectories, significantly improving fine-grained control, generalization, and resilience to perceptual or physical disturbances in grasping and pick-and-place tasks. Wu et al. \cite{wu2023learning} introduced a diffusion policy for grasping tasks and modeled grasping generation as a denoising process, demonstrating that this strategy is more effective at capturing multimodal grasping distributions. It laid the foundation for the subsequent grasping and pick-and-place task operations. DexDiffuser\cite{weng2024dexdiffuser} and GrainGrasp\cite{zhao2024graingrasp} considered the geometric and physical constraints of the object on this basis, enhancing the effect of contact operations. 

Additionally, GenDP\cite{wang2024gendp} starts from the semantics of functional grasping, enabling operations to go beyond merely relying on large-scale shape learning and to better understand the functions of objects and achieve generalization. SpringGrasp\cite{chen2024springgrasp} extends the grasping task of diffusion policy to flexible and variable objects, addressing the limitation that early operations were limited to rigid bodies. Pan et al.\cite{pan2024vision} combine VLA and diffusion policy to achieve advanced reasoning capabilities: VLA is responsible for identifying the grasping objects, and diffusion policy is responsible for handling the grasping strategies. Makarova et al.\cite{makarova2025diffusionrl} combine diffusion policy and RL, improving the sample efficiency of using a single method.

\textbf{Imitation Learning.} Imitation learning (IL) offers an effective alternative to reinforcement learning for grasping and pick-and-place tasks by directly leveraging expert demonstrations to learn fine-grained contact behaviors, while avoiding large-scale trial-and-error exploration~\cite{an2025dexterous}. Early vision-based approaches retarget human hand motions captured from videos to robotic hands~\cite{qin2022one}, and are later extended with 3D hand–object reconstruction to extract grasping trajectories from daily-life videos~\cite{qin2022dexmv}. To mitigate the morphological gap between human and robotic hands, adversarial domain adaptation further improves cross-instance generalization in video-based imitation learning~\cite{liu2025immimic}.

Beyond vision-only supervision, multimodal imitation integrates visual, tactile, and proprioceptive signals to improve grasp robustness under varying object properties~\cite{li2025softgrasp}, and proves particularly effective for tactile-dominant pick-and-place tasks such as paper picking~\cite{lin2025pp}. To address data efficiency, attention-based models enable one-shot or few-shot grasp learning~\cite{wang2024neural}, while reusable skill primitives support generalization to new objects and environments~\cite{li2025single, wei2023generalized}. In parallel, synthetic data pipelines such as CyberDemo scale supervision and demonstrate strong sim-to-real transfer~\cite{wang2024cyberdemo}.

\textbf{Representation Learning.} Representation learning provides stable, generalizable, and noise-robust visual features for grasping and pick-and-place, significantly improving policy learning efficiency, action planning, and cross-scene manipulation performance. D(R,O)~\cite{wei2024d} uses a distance matrix to encode the relationship between the hand’s geometry at a given grasp pose and the object’s shape. This representation-learning approach enables the model to learn a unified grasp representation across different hand embodiments, object categories, and configurations, allowing it to predict executable grasps directly from point clouds while remaining robust to partial or incomplete observations. UniDexGrasp~\cite{xu2023unidexgrasp} and UniDexGrasp++~\cite{wan2023unidexgrasp++} focus on learning universal dexterous grasping policies, with UniDexGrasp using object curriculum learning and DAgger distillation for category-level generalization, while UniDexGrasp++ incorporates geometry-aware representation learning for enhanced generalization across object shapes. Together, these models demonstrate that learning geometry-aware representations improves grasping generalization across hands, objects, and configurations. 

Furthermore, D3Grasp~\cite{wang2025d3grasp}  constructs a multimodal representation that fuses visual, tactile, and hand-object interaction states into a unified semantic feature, which is used to guide reinforcement learning strategies. D-Grasp~\cite{christen2022d} formulates dynamic hand–object interaction as a representation-learning problem: instead of operating in the raw state space, it learns object-centric and grasp-conditioned representations that, combined with physics simulation and reinforcement learning, enable the synthesis of stable, physically plausible grasp-to-motion trajectories. UniGraspTransformer~\cite{wang2025unigrasptransformer} uses a Transformer to learn a unified, generalizable grasp representation, improving cross-task and cross-object generalization. Offline distillation consolidates multiple expert policies into a single model, reducing data requirements while ensuring strong generalization and robustness.

\begin{figure}[h]
    \centering
    \includegraphics[width=1\linewidth]{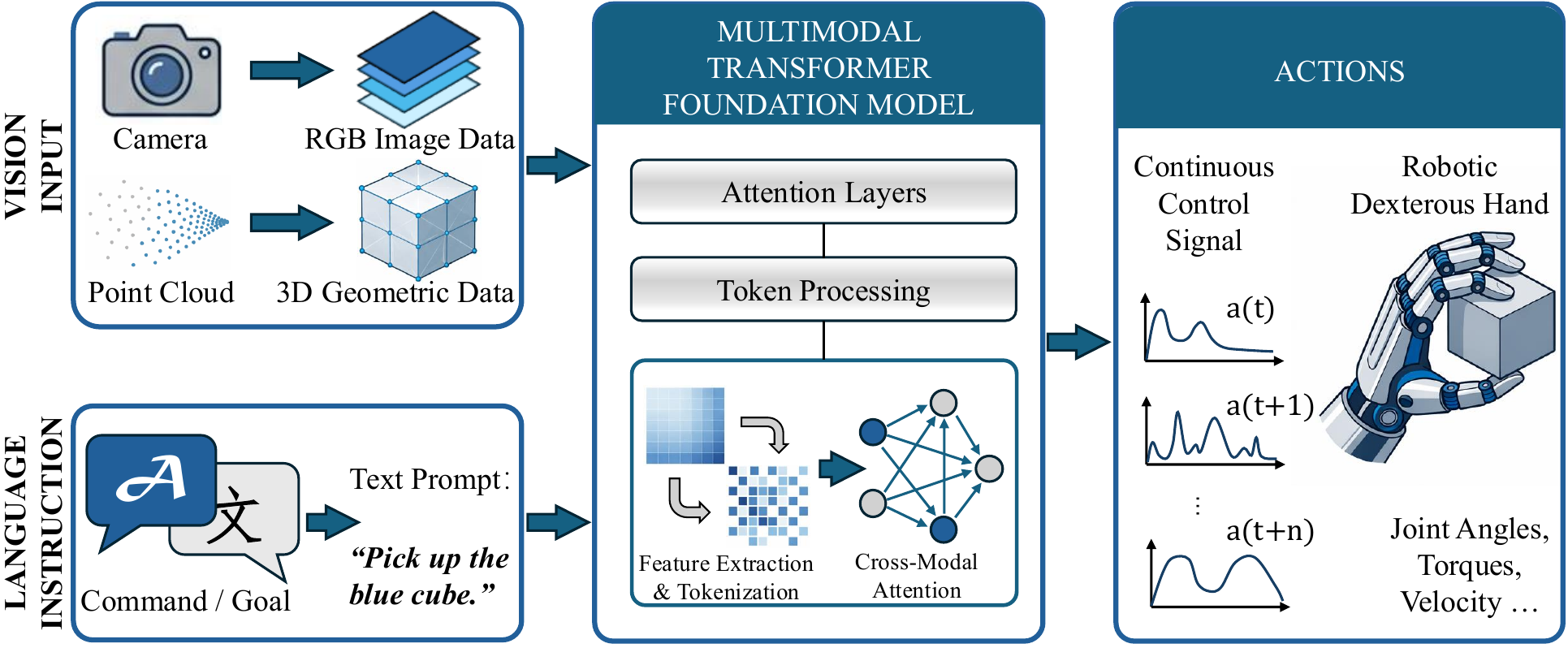}
    \caption{General architecture of a Vision-Language-Action (VLA) foundation model. Multimodal inputs (RGB/Depth and natural language instructions) are processed through a central Transformer model, which outputs low-level joint angle and torque commands for dexterous execution.}
    \label{fig:vla}
\end{figure}

\textbf{VLA.} As depicted in Fig.~\ref{fig:vla}, incorporating VLA models into grasping and pick-and-place enables semantic understanding, cross-object generalization, multi-task capability, and more natural human–robot interaction, thereby significantly improving the flexibility, robustness, and generality of manipulation policies. By integrating a VLA architecture with dexterous grasping, DexGraspVLA~\cite{zhong2025dexgraspvla} achieves unprecedented generalization and robustness in grasping and pick-and-place tasks. The system reaches over 90\% zero-shot grasp success across 1,287 unseen cluttered scenes, handles challenging objects such as transparent, reflective, and deformable items, and executes long-horizon language-conditioned tasks with an overall success rate of 89.6\%. It also supports non-prehensile pre-grasp strategies and remains robust under physical disturbances, positioning it as one of the most versatile and reliable VLA-driven grasping systems to date. 

Moreover, GraspVLA~\cite{deng2025graspvla} applies a unified Vision-Language-Action model to grasping and pick-and-place, achieving strong cross-object and cross-scene generalization. With SynGrasp-1B and large-scale semantic pretraining, it supports open-vocabulary understanding and zero-shot sim-to-real grasping, reaching $91.2\%$ success on unseen objects and showing robust performance under challenging real-world conditions. Pan et al. combine VLA-based semantic planning with diffusion policy–based fine manipulation, achieving the first language-conditioned, highly robust, and multimodal grasping system for dexterous multi-fingered pick-and-place\cite{pan2024vision}.

\textbf{Other.} Beyond the methods discussed above, several additional works have also demonstrated strong performance on grasping and pick-and-place tasks. Xiao et al.~\cite{xiao2025designing} use a deformable pin-pression gripper, together with geometry-aware representations and curriculum RL, to achieve robust dynamic grasping and online in-hand adjustment for complex objects, thereby improving adaptability and stability compared with traditional parallel-jaw grippers. DextrAH-RGB~\cite{singh2024dextrah} combines geometric fabrics, an RL-based teacher policy, and an RGB-only distilled student to form an end-to-end visuomotor control framework, achieving the first reliable sim-to-real transfer for RGB-based dexterous grasping and demonstrating strong robustness and generalization across diverse objects and real-world settings. Zhou et al.~\cite{zhou2023learning} uses external environmental cues and goal-conditioned RL to address occluded grasping, showing that extrinsic dexterity allows a simple gripper to execute complex grasp behaviors.

\subsection{Tool \& Device Operation}
\label{sec:tool}
\begin{figure}[ht]
    \centering
    \includegraphics[width=0.8\linewidth]{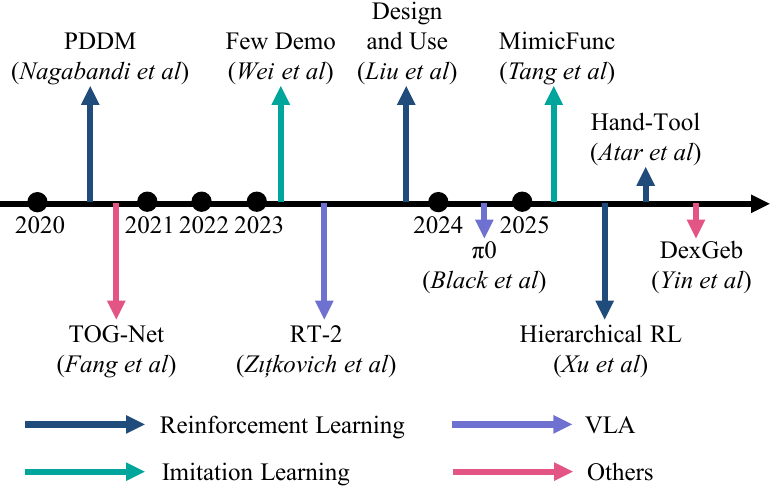}
    \caption{Tool Use \& Device Operation Timeline}
    \label{fig:tud}
\end{figure}

Tool and device operation represent some of the most challenging forms of dexterous manipulation, as they require reasoning about functional affordances rather than mere object geometry. Robots must understand how tools are intended to be used, adapt to variations in shape and mechanics, and generate precise force interactions to achieve task goals. Recent research explores multiple learning paradigms for this problem—from imitation learning that transfers human tool-use demonstrations, to reinforcement learning for mastering complex contact dynamics, to VLA models that inject semantic understanding of tool functions. As depicted in Fig.~\ref{fig:tud}, we review representative methods under each category. 

\textbf{Reinforcement Learning.} RL can learn complex contact dynamics, non-intuitive coordination strategies, and task-specific force behaviors in tool use and device operation, while also providing strong generalization and long-horizon planning abilities, making it a key approach for solving these challenging manipulation tasks. PDDM~\cite{nagabandi2020deep} combines deep dynamics models with online MPC, enabling RL to learn complex contact dynamics, fine-grained pose control, and non-intuitive manipulation strategies for tool use and device operation. This approach achieves high-precision dexterous manipulation in the real world with minimal data. Xu et al.~\cite{xu2025hierarchical} use hierarchical RL to decouple tool-deformation control from task-level planning, enabling a multifingered hand to learn complex three-way dynamics and non-intuitive tool-use strategies, ultimately achieving robust indirect grasping with articulated tools. Atar et al.~\cite{atar2025hand} employ sim-to-real reinforcement learning to master the joint dynamics of articulated tools, and use tactile-guided online adaptation to achieve stable operation and generalization to unseen tools. Liu et al.~\cite{liu2023learning} use a two-stage reinforcement learning framework that enables a robot to automatically design and deploy task-specific tools, leading to more efficient tool-use policy learning and stronger cross-task generalization.

\textbf{Imitation Learning.} A key challenge in tool-use tasks lies in the large intra-function variations of differences in shape, size, and category—which makes many approaches that rely on geometric or visual similarity struggle to achieve one-shot generalization. MimicFunc addresses the problem of enabling a robot to imitate tool manipulation from a single human video and generalize the skill to entirely novel tools. Its core IL mechanism lies in inferring functional correspondences from the single demonstration and using them to synthesize executable robot trajectories~\cite{tang2025mimicfunc}.Wei et al. propose a method that enables robots to learn functional grasping from only a few human demonstrations and generalize to completely novel tools with diverse shapes. From these minimal demonstrations, the system extracts functional keypoints of the tool, the key hand poses used by humans, and a distribution of grasping preferences. In doing so, the robot learns the functional structure of the tool rather than merely imitating its geometric appearance~\cite{wei2023generalized}.

\textbf{VLA.} VLA integrates visual perception with functional language semantics, enabling robots to understand tool purposes, decompose long-horizon tasks, generalize to new tools, and produce safer, more stable, and more precise tool-use behaviors through semantic alignment. RT-2~\cite{zitkovich2023rt} aligns visual inputs with large-scale language semantics, enabling robots to understand tool functions, select appropriate or alternative tools, and act correctly in unseen scenarios, thereby greatly improving generalization and decision-making in tool-use tasks. $\pi$0~\cite{black2024pi0visionlanguageactionflowmodel} introduces a VLA flow-based policy that aligns vision with language and produces continuous, high-frequency actions for multi-embodiment, cross-task manipulation. With large-scale pre-training and targeted post-training, it outperforms existing VLA and diffusion models, showing strong generalization and high performance across complex tasks.

\textbf{Other.} There are also several methods outside the categories above that achieve strong performance on this task. DexGen~\cite{yin2025dexteritygen} combines large-scale RL pretraining with a diffusion-based controller to build the first general-purpose dexterous hand foundation policy, transforming coarse high-level commands into safe, stable, and fine-grained motions, and enabling unprecedented real-world dexterity such as rotation, regrasping, and complex tool use.
Fang et al. use TOG-Net~\cite{fang2020learning} to jointly optimize task-oriented grasping and manipulation strategies, trained in a self-supervised learning framework, successfully achieving task-specific tool grasping and manipulation, providing stability and high generalization for complex tasks.

\subsection{Human Interaction}
\label{sec:human}
Human–robot interaction tasks involve continuous motion, uncertainty, and strict safety requirements, making policy learning more challenging than object-centric dexterity. Unlike purely mechanical manipulation, interaction requires understanding human intent, predicting changes in motion, and maintaining compliant contact. Recent research therefore focuses on learning interaction strategies from human behavior—either through imitation to inherit natural handover patterns, or through reinforcement learning to adapt safely in dynamic collaboration. Below, we summarize key developments in imitation learning and reinforcement learning for human interaction.

\textbf{Reinforcement Learning.} Reinforcement learning can handle the unpredictability of human motion, learn safe and compliant interaction strategies, and adapt to real-time feedback, making it a key approach for managing dynamic collaboration and contact uncertainty in human–robot interaction. Human-in-the-Loop Reinforcement Learning (HIL-RL)~\cite{luo2025precise} framework that enables RL to learn precise, stable interaction behaviors with human participation. Through preference-based RL, the policy is optimized to match human-preferred interaction styles rather than merely achieving task completion. By iteratively exploring and incorporating human corrections, RL compensates for missing demonstrations and significantly improves the robot’s adaptability to real human partners. Lin~\cite{lin2025sim} et al. enhance the reliability of RL in real-world interaction scenarios by combining structured rewards, interaction-aware visual representations, and robust sim-to-real transfer, enabling robots to adapt stably to external disturbances in dynamic, contact-rich bimanual tasks and significantly improving interaction capability.

\textbf{Imitation Learning.} In human interaction tasks, human motions are continuous, unpredictable, and subject to strict safety requirement. Imitation learning provides a direct prior on how humans perform handovers. In that case, imitation learning is a crucial method for human interaction tasks involving dexterous hands.
Christen et al. introduce a two-stage framework for learning human-to-robot handovers from point cloud input. Imitation learning is used in both stages. Firstly, using planner-generated expert demonstrations for pre-training in imitation learning. Then, finetuning with reinforcement learning in dynamic human-robot motion via a self-supervised teacher-student framework~\cite{christen2023learning}. VRB is a system that learns visual affordances from large-scale human interaction videos for deployment on real robots. It provides demonstration priors that are directly usable for imitation learning and VRB effectively addresses the common data scarcity problem in imitation learning by supplying large-scale, video guidance\cite{bahl2023affordances}.  In addition, several datasets can be used for interaction tasks within imitation-learning frameworks. OpenEgo is a large-scale egocentric human-interaction dataset whose core value lies in systematically capturing real human–object interaction processes, including 3D hand joints, fine-grained dexterous motions, bimanual coordination patterns, and intention-aligned action primitives. These signals provide the essential priors that enable robots to learn human interaction behaviors~\cite{jawaid2025openego}.

\subsection{Bimanual Manipulation}
\label{sec:bim}
\begin{figure}[ht]
    \centering
    \includegraphics[width=1\linewidth]{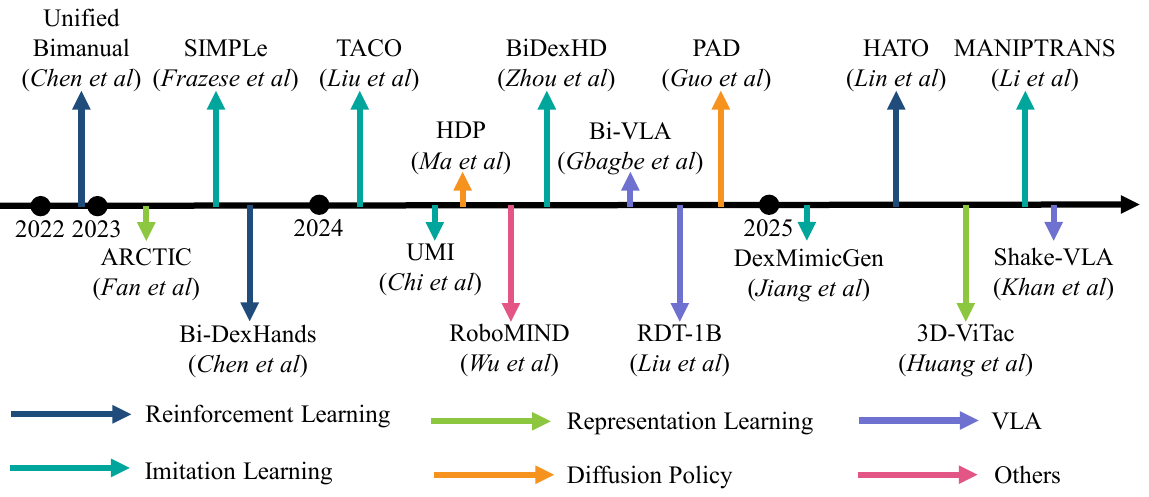}
    \caption{Bimanual Manipulation Timeline}
    \label{fig:bmt}
\end{figure}

Bimanual manipulation requires synchronized control of two high-DoF hands under complex contact dynamics, making coordination, force allocation, and stability notably more challenging than single-hand tasks. To address this, recent studies have developed dedicated learning frameworks and datasets that enable effective collaboration between both hands, as shown in Fig.~\ref{fig:bmt}. 

\textbf{Reinforcement Learning.} Reinforcement learning plays a crucial role in bimanual manipulation, enabling coordinated control in high-dimensional coupled systems, adapting to complex contact-rich dynamics, providing flexible online adjustment, and leveraging demonstrations to achieve strong generalization, thereby enhancing the robustness and intelligence of bimanual manipulation. Bi-DexHands\cite{chen2023bi} establishes the first systematic learning platform for bimanual dexterous manipulation, enabling RL to acquire multi-contact coordination and complex dynamic skills, and serving as a key benchmark for advancing human-level bimanual manipulation research. HATO~\cite{lin2025learning} presents the first scalable visuotactile imitation learning system that enables robots to learn real-world bimanual dexterous skills from human demonstrations and achieve high performance across multiple complex tasks. Chen et al.~\cite{chen2022towards} introduces a unified framework for bimanual dexterous manipulation that combines multimodal perception with IL–RL hybrid learning, enabling coordinated control, contact switching, and force allocation.

\textbf{Diffusion Policy.} Diffusion Policy can model multi-modal coordination strategies, generate smooth and coherent dual-arm trajectories, and remain robust with only a small number of demonstrations, making it an ideal approach for handling the high-dimensional coupling inherent in bimanual manipulation. In HDP~\cite{ma2024hierarchical}, the diffusion policy generates semantically aligned, physically feasible, and kinematically safe low-level control trajectories, making it a key component for achieving long-horizon and multi-task manipulation. PAD~\cite{guo2024prediction} unifies future visual prediction and action generation within a single diffusion denoising process, enabling the diffusion policy to learn more physically consistent and semantically aligned behaviors, which significantly improves generalization and real-world manipulation performance.

\textbf{Imitation Learning.} In bimanual manipulation tasks, which require tightly coordinated motions, human-like motion is difficult for reinforcement learning to discover. Imitation learning improves training efficiency, prevents unnatural behaviors, and enhances generalization.  Franzese et al. present SIMPLe, an Interactive Imitation Learning framework designed specifically for bimanual manipulation. By modeling demonstrations with a Graph Gaussian Process and allowing users to iteratively adjust trajectories, the method reduces the burden of demonstrations and enables dual-arm coordination to be learned through IL~\cite{franzese2023interactive}. Maniptrans~\cite{li2025maniptrans} frames dexterous bimanual manipulation as an imitation learning problem. A large human motion dataset is used to learn a human-like hand prior, and a small residual policy then adapts these imitated trajectories to satisfy physical constraints. This two-stage design based on IL enables the efficient transfer of human demonstrations to dexterous hands without task-specific rewards~\cite{li2025maniptrans}. BiDexHD uses a human motion dataset (e.g., TACO~\cite{liu2024taco}) and automatically parses trajectories, finger poses, and object poses.It converts each demonstration into a trainable task in the simulation. This removes the need for manually defining tasks and ensures that task diversity and quantity are no longer constrained~\cite{zhou2024learning}.

Furthermore, UMI is a pipeline that can transfer in-the-wild human demonstrations into a deployable dexterous hand policy. The approach eliminates the need for a physical robot during demonstrations, relying solely on a hand-held instrumented gripper to gather effective in-the-wild demonstrations for imitation learning, enabling the learned policies to zero-shot generalize when deployed on dexterous hands. It can achieve bimanual demonstrations via dual grippers~\cite{chi2024universal}. In addition, some models can generate bimanual manipulation data for imitation learning by expanding the training set through a real-to-sim pipeline. DexMimicGen introduces an automated data generation system for imitation learning in bimanual dexterous manipulation. It models three types of dual-arm subtasks—parallel, coordination, and sequential—and incorporates per-arm segmentation, asynchronous execution, synchronization, and ordering constraints. These mechanisms modularize otherwise tightly coupled bimanual actions and enable their automatic transformation and reproduction in simulation~\cite{jiang2025dexmimicgen}. 

\textbf{VLA.} VLA provides strong cross-object and cross-scene generalization in bimanual manipulation. It supports role allocation and coordination between the two hands, offers better task decomposition and error recovery for long-horizon operations, and unifies high-level semantic reasoning with low-level action generation, enabling interpretable high-level planning for complex bimanual tasks. Bi-VLA~\cite{gbagbe2024bi} builds a VLA system for bimanual dexterous manipulation, where visual perception interprets the scene, language models generate task plans, and action modules execute coordinated dual-arm behaviors, enabling multi-step tasks such as grasping, cutting, and mixing. The system executes complex bimanual tasks directly from language instructions without additional training. In RDT-1B~\cite{liu2024rdt}, VLA encodes language task objectives and visual scene semantics as conditioning inputs for the diffusion policy, enabling the model to generate coordinated and robust bimanual action sequences directly from natural language instructions. Shake-VLA~\cite{khan2025shake} introduces a Vision–Language–Action system for bimanual dexterous manipulation, specifically designed for multi-step cocktail-mixing tasks. The system integrates visual recognition, speech interaction, RAG-based recipe retrieval, anomaly detection, and action execution, enabling a complete closed-loop pipeline from natural language instructions to recipe lookup, ingredient verification, and coordinated dual-arm execution.

\textbf{Representation Learning.} Representation learning provides structured, semantic, and generalizable state spaces for bimanual manipulation, enabling policies to more easily acquire coordinated, multi-stage, and contact-rich behaviors while maintaining strong robustness across objects and environments. 3D-ViTac’s~\cite{huang20253dvitaclearningfinegrainedmanipulation} core contribution is the introduction of a unified 3D visuo-tactile representation that projects RGB-D point clouds and high-resolution tactile signals into a shared 3D coordinate space, producing a geometrically and semantically consistent multi-modal point-cloud embedding. This unified representation provides a structured, generalizable, and occlusion-robust state space, enabling diffusion policies to learn multi-contact, force-sensitive, and geometrically constrained bimanual manipulation more reliably. ARCTIC~\cite{fan2023arctic} provides a high-precision, temporally synchronized dataset of bimanual articulated hand–object interactions, enabling representation learning to be systematically trained for the first time in realistic, dynamic, multi-contact, and coordinated dual-hand manipulation scenarios.

\textbf{Other.} There are also several approaches that do not fall neatly into the categories above but nevertheless demonstrate strong performance on bimanual manipulation tasks. RoboMIND~\cite{wu2024robomind} offers a unified cross-embodiment, cross-task, multi-modal data foundation that enables representation learning to capture coordination patterns, contact dynamics, and semantic task structures in bimanual manipulation. This dataset provides essential support for the generalization of VLA, imitation learning, and diffusion policies in complex dual-arm tasks.

\begin{table*}[ht]
    \centering
    \scriptsize
    \setlength{\tabcolsep}{2pt}
    \renewcommand{\arraystretch}{1.05}
    \caption{Summary of major dexterous hand datasets.}
    \label{tab:dataset}
    
    \resizebox{\textwidth}{!}{%
    \begin{tabular}{@{} l l l l l l l @{}}
    \toprule
    \textbf{Dataset} & \textbf{Year} & \textbf{Scene} & \textbf{Modality} & \textbf{Embodiment} & \textbf{Collection} & \textbf{Scale} \\
    \midrule
    
    DexGraspNet~[181] & 2022 & Sim. & Pose, Object, Point cloud & Shadow Hand & Optimization-based & 1.32M poses; 5355 objects \\
    GenDexGrasp~[182] & 2022 & Sim. & Pose, Contact, Object & \makecell[l]{Barrett; Robotiq; Allegro; Shadow Hand} & Optimization-based & 436K poses; 58 objects \\
    DexMV~\cite{qin2022dexmv} & 2022 & Real+Sim. & \makecell[l]{RGB, Pose, Point cloud} & Adroit Hand & Human capture & \makecell[l]{700 demos;} \\
    DexArt~\cite{bao2023dexart} & 2023 & Sim. & Point cloud, Proprioception & \makecell[l]{Adroit; Allegro} & learning-based & 82 objects \\
    UniDexGrasp++~\cite{wan2023unidexgrasp++} & 2023 & Sim. & Pose, Point cloud, Proprioception & Shadow Hand & learning-based & \makecell[l]{1.32M poses; 5355 objects} \\
    RealDex~\cite{liu2024realdex} & 2024 & Real & Pose, RGB-D, Point cloud & Shadow Hand & Human capture & \makecell[l]{59K poses; 52 objects; 2.6K seqs.} \\
    DexCap~\cite{wang2024dexcap} & 2024 & Real & Pose, RGB-D, Point cloud & LEAP Hand & Human capture & 3.5h demo \\
    DexFuncGrasp~\cite{hang2024dexfuncgrasp} & 2024 & Real+Sim. & Pose, RGB-D, Contact & Shadow Hand & Human capture & \makecell[l]{14K+  Pose; 559 objects} \\
    RH20T~\cite{fang2024rh20t} & 2024 & Real & \makecell[l]{RGB-D; Contact; Proprioception} & \makecell[l]{Franka; Allegro;} & Human capture & \makecell[l]{110K+ seqs.} \\
    DexGraspAnything~\cite{zhong2025dexgrasp} & 2025 & Sim. & \makecell[l]{Pose, RGB-D, Point cloud, Semantics} & Shadow Hand & learning-based & 3.4M grasps; 15.7K objects \\
    BODex~\cite{chen2025bodex} & 2025 & Sim. & Pose, Contact, Object & Shadow Hand & Optimization-based & 3.08M grasps; 2397 objects \\
    CEDex~\cite{wu2025cedex} & 2025 & Sim. & Pose, Contact, Point cloud & \makecell[l]{Allegro; LEAP Hand} & learning-based & 20M grasps; 500K objects \\
    Dex1B~\cite{ye2025dex1b} & 2025 & Sim. & Pose, Contact, Object & \makecell[l]{Inspire; Ability Hand} & learning-based & 1B demos; 6K+ objects \\
    DexTOG~\cite{Zhang2025dextog} & 2025 & Sim. & Pose, Point cloud, Semantics & Shadow Hand & learning-based & 80K grasps; 80 objects\\
    VTDexManip~\cite{liu2025vtdexmanip} & 2025 & Sim. & RGB, Contact & Shadow Hand & Human capture &182 objects;  2032 seqs. \\
    
    \bottomrule
    \end{tabular}%
    }
    \vspace{-1.0em}
    \end{table*}

\section{Dataset of Dexterous Hand Research}
\label{sec:dataset_eval}
This section surveys representative datasets for dexterous-hand research from three interconnected perspectives: data acquisition (Section~\ref{sec:acqu}), modality design (Section~\ref{subsec:dataset_modalities}), and evaluation (Section~\ref{subsec:dataset_metrics}).  Furthermore, the overview of representative datasets is summarized in Table~\ref{tab:dataset}.

\subsection{Data Acquisition}
\label{sec:acqu}
From a collection-method perspective, recent dexterous-hand datasets can be grouped into three families, summarized in the Collection column of Table~\ref{tab:dataset}: optimization-based acquisition in simulation, learning-based acquisition in simulation, and human capture. The distinction is whether labels come mainly from numerical optimization, from learned or generative models, or from human demonstrations and wearable sensing in the real world, optionally followed by simulation-based verification or scaling.

\textbf{Optimization-based.} These pipelines synthesize grasps or interaction poses by minimizing analytic or differentiable objectives, such as force-closure energies, bilevel programs with QP subproblems, or SDF-based refinement, and they validate candidates in physics engines such as MuJoCo or Isaac Gym~\cite{todorov2012mujoco}. They offer high throughput and reproducible labels, but quality still depends on energy design and the sim-to-real gap.

\textbf{Learning-based.} These~\cite{bao2023dexart,wan2023unidexgrasp++,wu2025cedex,ye2025dex1b} data are produced or enlarged by learned components: reinforcement learning with curricula and distillation, diffusion or other generative models, retrieval-based and rule-guided data engines, or model-in-the-loop filtering and expansion. Such methods can reach very large scales and encode complex distributions, at the cost of training stability, compute, and careful alignment with physical plausibility.

\textbf{Human Capture.} This family records humans or teleoperated robots directly, using multi-view RGB-D, wearable motion capture, electromagnetic or inertial gloves, tactile arrays, or similar modalities~\cite{garcia2018first, brahmbhatt2020contactpose, chao2021dexycb, yang2022oakink}. It yields realistic contact and behavior priors but can be expensive to scale. Many systems combine human acquisition with downstream simulation; for example, retargeting human motions to dexterous hands and checking grasps in sim, or augmenting mocap with reinforcement learning rollouts. In those cases, the boundary to pure simulation is methodological rather than exclusive.

Broader robot-learning efforts also aggregate heterogeneous datasets for cross-embodiment training~\cite{o2024open}, and multimodal capture systems continue to blend sensing modalities to improve coverage. These trends complement the three collection families above.

\subsection{Data Modality}
\label{subsec:dataset_modalities}

The Modalities column in Table II shows that curated dexterous manipulation datasets are increasingly organized around complementary observation channels rather than a single sensing stream. The most common foundation is geometric state, including hand and object poses, object meshes or point clouds, and robot proprioception. These signals provide the structural information required for grasp synthesis, state estimation, and object-centric control. In parallel, visual observations such as RGB or RGB-D are widely adopted~\cite{qin2022dexmv,liu2024realdex,wang2024dexcap,hang2024dexfuncgrasp,fang2024rh20t,zhong2025dexgrasp,liu2025vtdexmanip}, especially in real-world and hybrid pipelines, because they preserve scene appearance, occlusion, and cross-instance variation that geometry alone cannot capture. Geometry and visual appearance form the perceptual basis of most current dexterous learning pipelines.

A second trend is the growing inclusion of contact-related modalities, reflecting the fundamentally contact-rich nature of dexterous manipulation. Beyond pose and vision, many recent datasets~\cite{li2022gendexgrasp,hang2024dexfuncgrasp,fang2024rh20t,chen2025bodex,wu2025cedex,ye2025dex1b,liu2025vtdexmanip} explicitly record contact labels, tactile signals, force and torque measurements, or contact-aware visual observations. These modalities are particularly important for grasp stability, slip detection, force regulation, and fine manipulation under partial observability. Thus, curated datasets differ not only in scale, but also in how contact is represented: some remain geometry-dominant and mainly support grasp generation, whereas others incorporate richer interaction signals and are better suited for closed-loop manipulation and tactile-reactive control.

Another emerging direction is the integration of semantics and task-level conditioning. Several recent datasets~\cite{zhong2025dexgrasp,Zhang2025dextog} augment geometric and contact-centric observations with semantic labels, task descriptors, or language-level annotations, enabling learning beyond isolated grasp generation toward task-aware and long-horizon manipulation. This shift matters because dexterous behavior is rarely defined only by contact feasibility; it also depends on what action should be performed, under which object function, and toward which downstream goal. Therefore, dataset modalities should be understood not merely as sensing choices, but also as assumptions about the level of manipulation intelligence a benchmark can support. In practice, modality composition largely determines whether a dataset is best suited for grasp prediction, contact-rich closed-loop control, or semantically conditioned manipulation.

\subsection{Evaluation}
\label{subsec:dataset_metrics}

\textbf{Evaluation Metric.} Evaluation in curated dexterous-hand datasets is highly heterogeneous and cannot be reduced to a single notion of success. In practice, existing works usually assess at least two layers of performance: the quality of grasps or poses prior to execution, and the performance of policies or generators during downstream execution. For grasp- and pose-centric datasets, the most common criteria concern physical plausibility, including penetration, analytic or quasi-static stability, and diversity over generated grasp proposals or object geometries. Representative examples include DexGraspNet, GenDexGrasp, CeDex, BODex, and DexGrasp, which emphasize whether synthesized grasps are physically feasible and sufficiently diverse before policy deployment~\cite{wang2022dexgraspnet,li2022gendexgrasp,wu2025cedex,chen2025bodex,zhong2025dexgrasp}. Related benchmarks such as DexTOG further connect grasp-level quality with downstream utility by jointly considering grasp validity and policy performance, explicitly distinguishing analytic grasp quality from executable manipulation value~\cite{Zhang2025dextog}.

A second evaluation dimension concerns whether policies trained on a dataset can reliably execute manipulation tasks and generalize beyond the conditions seen during training. Here, task success rate remains the dominant endpoint, but its interpretation depends strongly on the evaluation protocol. Simulation-oriented benchmarks such as DexArt, UniDexGrasp++, and Dex1B evaluate not only nominal task completion, but also transfer across held-out object instances, unseen categories, or broader task settings~\cite{bao2023dexart,wan2023unidexgrasp++,ye2025dex1b}. Datasets with tactile sensing introduce an additional robustness axis. For example, VTDexManip emphasizes performance under contact uncertainty and evaluates manipulation under variations such as unseen objects, viewpoint changes, tactile thresholds, and tactile noise~\cite{liu2025vtdexmanip}. Taken together, these works show that the success rate is meaningful only when paired with a clear specification of whether it reflects nominal execution, cross-instance generalization, or robustness to sensing perturbations~\cite{bao2023dexart,wan2023unidexgrasp++,liu2025vtdexmanip,ye2025dex1b}.

For real-world or human-capture datasets, reporting often shifts from grasp validity alone toward motion fidelity, stage completion, and demonstration quality. RH20T, for instance, supports long-horizon human manipulation analysis and is commonly associated with stage-aware or temporally structured evaluation rather than only single-step task completion~\cite{fang2024rh20t}. DexCap and related imitation-oriented settings place greater emphasis on repeated-trial task success, subtask completion, and data efficiency, making partial progress distinguishable from full execution success~\cite{wang2024dexcap}. Other real or hybrid datasets, including RealDex, DexMV, and DexFuncGrasp, additionally emphasize real-robot success, trajectory consistency, contact realism, or motion reconstruction fidelity, reflecting that real-world dexterous manipulation is often evaluated not only by task completion, but also by how stably, naturally, and efficiently the behavior is executed~\cite{liu2024realdex,qin2022dexmv,hang2024dexfuncgrasp}. These protocols indicate that current reporting practices combine multiple notions of quality, including grasp validity, execution success, generalization, robustness, motion realism, and demonstration efficiency~\cite{liu2024realdex,wang2024dexcap,fang2024rh20t,qin2022dexmv,hang2024dexfuncgrasp}.

\textbf{Reliability and Trustworthiness.} Beyond conventional performance metrics, reliability and trustworthiness are increasingly important for dexterous hand systems, especially when learning-based policies are deployed in open, contact-rich, and human-centered environments~\cite{Intelligence202505AV}. Real-world deployment requires not only high task success but also reliable perception, robust decision making, safe interaction, and predictable failure behavior. A system may perform well under nominal benchmark settings, yet still fail under occlusion, sensor noise, ambiguous object boundaries, unseen objects, changing contact conditions, or human intervention~\cite{bai2025retrdex}. Therefore, evaluation protocols should move beyond benchmark-centered assessment and include reliability-oriented criteria.

A key requirement is to evaluate whether the system can perceive, decide, and recover reliably under uncertainty and distribution shift. Dexterous hand systems depend on object detection, segmentation, pose estimation, affordance prediction, and contact-region understanding, where errors may propagate to grasp planning, force control, and policy execution. Future evaluation should therefore assess not only perception accuracy and task success, but also uncertainty calibration, robustness to occlusion and clutter, out-of-distribution detection, failure prediction, recovery rate, safety violation rate, and multimodal consistency across vision, tactile sensing, and proprioception. These criteria can better reveal whether a dexterous hand system is not only capable, but also reliable, interpretable, and safe in open environments~\cite{Fang2024MOKAOR,zhao20243d,Li2025HAMSTERHA}.

\section{Current Limitations and Future Directions}
\label{sec:challenge_future}
Although dexterous hand research has made remarkable progress in recent years, it still faces several fundamental challenges that limit the applicability of existing technologies and, at the same time, point to the most critical future directions. Rather than discussing limitations and future work separately, it is more coherent to connect each current bottleneck with its corresponding research avenue.

\textbf{From the trade-off between bionic complexity and hardware feasibility to scalable dexterous hand design.} The design of dexterous hands faces a fundamental trade-off between high degrees of freedom and system complexity. Achieving human-like manipulation requires tightly integrated actuation, perception, and control, which increases mechanical complexity, computational demand, and system cost. As a result, existing robotic hands either simplify structure at the expense of manipulation capability or suffer from limited reliability, scalability, and maintainability due to highly integrated mechatronic systems. A promising direction is to move from mechanical complexity toward functional intelligence. Rather than merely mimicking appearance or structure, future research should focus more on the biomechanical and neural control mechanisms of the human hand~\cite{rothemund2021hasel}. In this regard, motion synergies offer an effective means to reduce control dimensionality, allowing coordinated movements of multiple joints to be driven by a single control signal.

\textbf{From the dilemma of multimodal perception fusion to robust and interpretable perception.} Despite increasing sensor integration, effective multimodal perception fusion in dexterous hands remains challenging. Differences in sensing modalities, update rates, noise characteristics, and spatial locality make real-time synchronization and interpretation difficult. As a result, achieving stable, low-latency perception--control loops under contact-rich conditions remains a major bottleneck for human-level dexterity. To address this, perception systems should move beyond simple multi-sensor integration toward cross-modal contextual understanding, with stronger emphasis on robustness, generalization, and trustworthiness under noisy or partially missing sensory inputs.

\textbf{From limitations in learning, control, and data to robust and generalizable learning frameworks beyond benchmark-centric optimization.} Traditional model-based control relies on accurate dynamics models, yet the complex and discontinuous contact interactions in dexterous manipulation make precise modeling difficult, leading to performance degradation. Although data-driven methods have become dominant, learning-based approaches still suffer from distribution shift, limited generalization, safety concerns, and heavy computational demands, especially when combined with large models~\cite{vemprala2024chatgpt}. These challenges are further aggravated by data limitations. Existing datasets also lack sufficient diversity in long-horizon interactions, failure cases, and recovery behaviors, and tactile information is often underutilized. Looking forward, learning frameworks should move beyond benchmark-centric optimization toward robust, generalizable, and trustworthy policy learning~\cite{zhang2023survey,zhang2024survey}, so that commands can be interpreted and executed in physically plausible and semantically meaningful ways~\cite{brohan2022rt,vemprala2024chatgpt}.

\textbf{From system integration and industrialization barriers to deployable dexterous platforms.} Cost and reliability remain the primary barriers to the industrial deployment of dexterous hands. High-performance systems such as the Shadow Hand are expensive to manufacture and maintain, while low-cost or open-source alternatives often sacrifice sensing quality, material durability, or long-term reliability, limiting large-scale adoption~\cite{andrychowicz2020learning}. These issues are further amplified by the high training and deployment cost of dexterous manipulation policies. High-dimensional state--action spaces impose substantial sample and computational demands, and the sim-to-real gap continues to degrade real-world performance despite domain randomization and adaptation efforts~\cite{zitkovich2023rt,zhao2025high,hao2025learn}. In addition, the absence of unified performance evaluation standards and benchmark protocols hampers objective comparison across systems, slowing standardization and industrialization. Establishing authoritative evaluation frameworks is therefore essential for the sustainable development of dexterous hand technologies~\cite{huang2025human}. A practical future direction is to move from laboratory prototypes toward deployable dexterous platforms through modular system design, standardized interfaces, and plug-and-play components such as fingers, joints, and sensors, thereby making manufacturing, maintenance, and upgrades more affordable and scalable.

\section{Conclusion}

This survey revisits dexterous hand research from a holistic perspective and organizes the field around hardware design, methodological development in control and learning, datasets and evaluation practices, and the field's major limitations and future directions. At the hardware level, we summarize key developments in actuation, transmission, perception, and representative hand designs, showing that dexterous hand design is fundamentally shaped by trade-offs in force capability, compliance, bandwidth, integration, and system complexity. At the methodological level, we review the evolution of control and learning approaches, highlighting how different paradigms have emerged to address increasingly complex dexterous manipulation problems. We further consolidate datasets, modality design, and evaluation practices, emphasizing that methodological progress should be interpreted together with the ways in which dexterous manipulation is trained, benchmarked, and assessed.

Despite substantial progress, dexterous hand research still faces several fundamental challenges. These include the trade-off between dexterity and hardware complexity, the difficulty of integrating and synchronizing multimodal perception, the limited generalization of learning-based methods, the constraints imposed by data quality and scale, and the practical barriers of system integration and industrial deployment. In addition, the lack of unified evaluation standards continues to hinder fair comparison across different systems and methods. Looking forward, further progress will depend on more effective hardware design, better integration of sensing and control, more scalable and informative datasets, more transferable and data-efficient learning methods, and more standardized evaluation frameworks. We hope this survey provides a structured framework for understanding the development of dexterous hands and clarifies promising directions for future research.

\bibliographystyle{ieeetr}
\bibliography{reference.bib}

@String(IROS = {IEEE/RSJ International Conference on Intelligent Robots and Systems (IROS)})

@String(ICRA = {IEEE International Conference on Robotics and Automation (ICRA)})

@String(CORL = {Conference on Robot Learning (CoRL)})

@String(RSS = {Robotics: Science and Systems (RSS)})

@String(Humanoids = {IEEE/RAS International Conference on Humanoid Robots (Humanoids)})

@String(ROBIO = {IEEE International Conference on Robotics and Biomimetics (ROBIO)})

@String(CVPR = {IEEE/CVF conference on Computer Vision and Pattern Recognition (CVPR)})

@String(ICCV = {IEEE/CVF International Conference on Computer Vision (ICCV)})

@String(ECCV = {European Conference on Computer Vision (ECCV)})

@String(NeurIPS = {Advances in Neural Information Processing Systems (NeurIPS)})

@String(ICML = {International Conference on Machine Learning (ICML)})

@String(AAAI = {AAAI Conference on Artificial Intelligence (AAAI)})

@String(ICLR = {International Conference on Learning Representations (ICLR)})

@String(TPAMI = {IEEE Transactions on Pattern Analysis and Machine Intelligence (TPAMI)})

@String(TOG = {ACM Transactions on Graphics (TOG)})

@String(EAAI = {Engineering Applications of Artificial Intelligence (EAAI)})

@String(SciAdv = {Science Advances})

@String(NANO = {ACS Nano})

@String(IJRR = {The International Journal of Robotics Research})

@String(TROA = {IEEE Transactions on Robotics and Automation})

@String(TMECH = {IEEE/ASME Transactions on Mechatronics (T-MECH)})

@String(TRO = {IEEE Transactions on Robotics (T-RO)})

@String(RCIM = {Robotics and Computer-Integrated Manufacturing (RCIM)})

@String(TACON = {IEEE Transactions on Automatic Control (T-ACON)})

@String(TMRB = {IEEE Transactions on Medical Robotics and Bionics (T-MRB)})

@String(TASE = {IEEE Transactions on Automation Science and Engineering (T-ASE)})

@String(RA-L = {IEEE Robotics and Automation Letters (RA-L)})

@ARTICLE{bicchi2000dexteroushand,
  author={Bicchi, A.},
  journal=TROA, 
  title={Hands for dexterous manipulation and robust grasping: a difficult road toward simplicity}, 
  year={2000},
  volume={16},
  number={6},
  pages={652-662},
}

@article{vertongen2020mechanical,
  title={Mechanical aspects of robot hands, active hand orthoses, and prostheses: A comparative review},
  author={Vertongen, Jens and Kamper, Derek G and Smit, Gerwin and Vallery, Heike},
  journal=TMECH,
  volume={26},
  number={2},
  pages={955--965},
  year={2020},
}

@article{pozzi2023actuated,
  title={Actuated palms for soft robotic hands: review and perspectives},
  author={Pozzi, Maria and Malvezzi, Monica and Prattichizzo, Domenico and Salvietti, Gionata},
  journal=TMECH,
  volume={29},
  number={2},
  pages={902--912},
  year={2023},
}

@article{gu2023soft,
  title={Soft robotics enables neuroprosthetic hand design},
  author={Gu, Guoying and Zhang, Ningbin and Chen, Chen and Xu, Haipeng and Zhu, Xiangyang},
  journal=NANO,
  volume={17},
  number={11},
  pages={9661--9672},
  year={2023},
}

@article{yousef2011tactile,
  title        = {Tactile sensing for dexterous in-hand manipulation in robotics---A review},
  author       = {Yousef, Hanna and Boukallel, Mehdi and Althoefer, Kaspar},
  journal      = {Sensors and Actuators A: Physical},
  volume       = {167},
  number       = {2},
  pages        = {171--187},
  year         = {2011},

}

@article{li2020review,
  title={A review of tactile information: Perception and action through touch},
  author={Li, Qiang and Kroemer, Oliver and Su, Zhe and Veiga, Filipe Fernandes and Kaboli, Mohsen and Ritter, Helge Joachim},
  journal=TRO,
  volume={36},
  number={6},
  pages={1619--1634},
  year={2020},
}

@article{liang2025structure,
  title={The structure, material and performance of multi-functional tactile sensor and its application in robot field: A review},
  author={Liang, Tianwei and Liu, Zirui and Zhang, Hao and Zhou, Xue and Liang, Yunhong},
  journal={Materials Today},
  year={2025},
}

@article{zhou2022non,
  title={Non-invasive human-machine interface (HMI) systems with hybrid on-body sensors for controlling upper-limb prosthesis: A review},
  author={Zhou, Hao and Alici, Gursel},
  journal={IEEE Sensors Journal},
  volume={22},
  number={11},
  pages={10292--10307},
  year={2022},
}

@article{song2025Learning,
  title   = {An Overview of Learning-Based Dexterous Grasping: Recent Advances and Future Directions},
  author  = {Song, X. and Li, Y. and Zhang, Y. and others},
  journal = {Artificial Intelligence Review},
  volume  = {58},
  pages   = {300},
  year    = {2025},
}

@article{an2025dexterous,
  title={Dexterous manipulation through imitation learning: A survey},
  author={An, Shan and Meng, Ziyu and Tang, Chao and Zhou, Yuning and Liu, Tengyu and Ding, Fangqiang and Zhang, Shufang and Mu, Yao and Song, Ran and Zhang, Wei and others},
  journal={arXiv preprint arXiv:2504.03515},
  year={2025}
}

@article{huang2025human,
  title={Human-like dexterous manipulation for anthropomorphic five-fingered hands: A review},
  author={Huang, Yayu and Fan, Dongxuan and Duan, Haonan and Yan, Dashun and Qi, Wen and Sun, Jia and Liu, Qian and Wang, Peng},
  journal={Biomimetic Intelligence and Robotics},
  pages={100212},
  year={2025},
}

@misc{welte2025interact,
      title={Interactive Imitation Learning for Dexterous Robotic Manipulation: Challenges and Perspectives -- A Survey}, 
      author={Edgar Welte and Rania Rayyes},
      year={2025},
      eprint={2506.00098},
      archivePrefix={arXiv},
}

@article{firth2022anthropomorphic,
  title={Anthropomorphic soft robotic end-effector for use with collaborative robots in the construction industry},
  author={Firth, Charlotte and Dunn, Kate and Haeusler, M Hank and Sun, Yi},
  journal={Automation in Construction},
  volume={138},
  pages={104218},
  year={2022},
}

@article{wang2025towards,
  title={Towards Damage-Less Robotic Fragile Fruit Grasping: A Systematic Review on System Design, End Effector, and Visual and Tactile Feedback},
  author={Wang, Qingyu and Tu, Yuyang and Xu, Weidong and Zhang, Jianwei and Knoll, Alois and Zhou, Mingchuan and Ying, Yibin},
  journal={Journal of Field Robotics},
  year={2025},
}

@article{gao2025empower,
  title={Empower dexterous robotic hand for human-centric smart manufacturing: A perception and skill learning perspective},
  author={Gao, Benhua and Fan, Junming and Zheng, Pai},
  journal=RCIM,
  volume={93},
  pages={102909},
  year={2025},
}

@article{zhang2025dexterous,
title = {Dexterous hand towards intelligent manufacturing: A review of technologies, trends, and potential applications},
journal = RCIM,
volume = {95},
pages = {103021},
year = {2025},
issn = {0736-5845},
doi = {https://doi.org/10.1016/j.rcim.2025.103021},
author = {Jiexin Zhang and Huan Zhao and Kuangda Chen and Guanbo Fei and Xiangfei Li and Yiwei Wang and Zeyuan Yang and Shengwei Zheng and Shiqi Liu and Han Ding},

}

@ARTICLE{skaar2007def,      
  author={Skaar, Steven B.},
  journal=TACON, 
  title={Robot Modeling and Control-[Book review; M. Spong, S. Hutchinson, and M. Vidyasagar]}, 
  year={2007},
  volume={52},
  number={2},
  pages={378-379},
}

@inproceedings{zhang2018magneto,
  title={The magneto-thermal analysis of a high torque density joint motor for humanoid robots},
  author={Zhang, Wu and Yu, Zhangguo and Chen, Xuechao and Huang, Qiang},
  booktitle=Humanoids,
  pages={112--117},
  year={2018},
}

@article{guo2024design,
  title={Design and analysis of a permanent magnet frameless motor},
  author={Guo, Kaikai and Guo, Youguang and Fang, Shuhua and Li, Cong and Xue, Wangqi},
  journal={IEEE Journal of Emerging and Selected Topics in Power Electronics},
  volume={12},
  number={3},
  pages={3124--3134},
  year={2024},
}

@article{guo2023simulation,
  title={Simulation Analysis of a Sandwich Cantilever Ultrasonic Motor for a Dexterous Prosthetic Hand},
  author={Guo, Kai and Lu, Jingxin and Yang, Hongbo},
  journal={Micromachines},
  volume={14},
  number={12},
  pages={2150},
  year={2023},
}

@inproceedings{izuhara2018miniature,
  title={Miniature robot finger using a micro linear ultrasonic motor and a closed-loop linkage},
  author={Izuhara, Shunsuke and Mashimo, Tomoaki},
  booktitle=IROS,
  pages={1--9},
  year={2018},
}

@article{kargov2008development,
  title={Development of a miniaturised hydraulic actuation system for artificial hands},
  author={Kargov, A and Werner, Tino and Pylatiuk, Christian and Schulz, Stefan},
  journal={Sensors and Actuators A: Physical},
  volume={141},
  number={2},
  pages={548--557},
  year={2008},
}

@article{zhou2024dexterous,
  title={A dexterous and compliant hand based on soft hydraulic actuation for human inspired fine in-hand manipulation},
  author={Zhou, Jianshu and Huang, Junda and Dou, Qi and Abeel, Pieter and Liu, Yunhui},
  journal=TRO,
  year={2024},
}

@article{chou1996measurement,
  title={Measurement and modeling of McKibben pneumatic artificial muscles},
  author={Chou, Ching-Ping and Hannaford, Blake},
  journal=TROA,
  volume={12},
  number={1},
  pages={90--102},
  year={1996},
}

@inproceedings{bishop2013force,
  title={Force and moment generation of fiber-reinforced pneumatic soft actuators},
  author={Bishop-Moser, Joshua and Krishnan, Girish and Kota, Sridhar},
  booktitle=IROS,
  pages={4460--4465},
  year={2013},
}

@article{hadi2023programmable,
  title={Programmable soft robotics actuator with Pneumatic Networks (pneunets)},
  author={Hadi, Yulyan Wahyu and Sunarya, Basilius Agung Yason and Alifdhyatra, Athar Fadlankahlil and Hidayat, Egi and Salomo, Jonathan and Purwidyantri, Agnes and Prabowo, Briliant Adhi and Anshori, Isa},
  journal={IEEE Sensors Journal},
  volume={23},
  number={17},
  pages={19382--19389},
  year={2023},
}

@article{robertson2017new,
  title={New soft robots really suck: Vacuum-powered systems empower diverse capabilities},
  author={Robertson, Matthew A and Paik, Jamie},
  journal={Science Robotics},
  volume={2},
  number={9},
  pages={eaan6357},
  year={2017},
}

@article{lee2024design,
  title={Design and analysis of reconfigurable origami-based vacuum pneumatic artificial muscles for versatile robotic system},
  author={Lee, Jin-Gyu and Rodrigue, Hugo},
  journal={Soft Robotics},
  volume={11},
  number={6},
  pages={984--993},
  year={2024},
}

@article{tawk20193d,
  title={3D printable linear soft vacuum actuators: their modeling, performance quantification and application in soft robotic systems},
  author={Tawk, Charbel and Spinks, Geoffrey M and in het Panhuis, Marc and Alici, Gursel},
  journal=TMECH,
  volume={24},
  number={5},
  pages={2118--2129},
  year={2019},
}

@article{baek2023dexterous,
  title={Dexterous robotic hand based on rotational shape memory alloy actuator-joints},
  author={Baek, Hangyeol and Khan, Abdul Manan and Bijalwan, Vishwanath and Jeon, Sangmin and Kim, Youngshik},
  journal=TMRB,
  volume={5},
  number={4},
  pages={1082--1092},
  year={2023},
}

@article{brochu2010advances,
  title={Advances in dielectric elastomers for actuators and artificial muscles},
  author={Brochu, Paul and Pei, Qibing},
  journal={Macromolecular Rapid Communications},
  volume={31},
  number={1},
  pages={10--36},
  year={2010},
}

@article{zhang2025review,
  title={A review of the applications and challenges of dielectric elastomer actuators in soft robotics},
  author={Zhang, Qinghai and Yu, Wei and Zhao, Jianghua and Meng, Chuizhou and Guo, Shijie},
  journal={Machines},
  volume={13},
  number={2},
  pages={101},
  year={2025},
}

@article{seong2024multifunctional,
  title={Multifunctional magnetic muscles for soft robotics},
  author={Seong, Minho and Sun, Kahyun and Kim, Somi and Kwon, Hyukjoo and Lee, Sang-Woo and Veerla, Sarath Chandra and Kang, Dong Kwan and Kim, Jaeil and Kondaveeti, Stalin and Tawfik, Salah M and others},
  journal={Nature Communications (NC)},
  volume={15},
  number={1},
  pages={7929},
  year={2024},
}

@article{zhang2023piezo,
  title={Piezo robotic hand for motion manipulation from micro to macro},
  author={Zhang, Shijing and Liu, Yingxiang and Deng, Jie and Gao, Xiang and Li, Jing and Wang, Weiyi and Xun, Mingxin and Ma, Xuefeng and Chang, Qingbing and Liu, Junkao and others},
  journal={Nature Communications (NC)},
  volume={14},
  number={1},
  pages={500},
  year={2023},
}

@article{zhang2025biomimetic,
  title={Biomimetic rigid-soft finger design for highly dexterous and adaptive robotic hands},
  author={Zhang, Ningbin and Zhou, Peiwei and Yang, Xinyu and Shen, Fengjie and Ren, Jieji and Hou, Tengyu and Dong, Le and Bian, Rong and Wang, Dong and Gu, Guoying and others},
  journal=SciAdv,
  volume={11},
  number={17},
  year={2025},
}

@article{yang2025lightweight,
  title={A lightweight prosthetic hand with 19-DOF dexterity and human-level functions},
  author={Yang, Hao and Tao, Zhe and Yang, Jian and Ma, Wenpeng and Zhang, Haoyu and Xu, Min and Wu, Ming and Sun, Shuaishuai and Jin, Hu and Li, Weihua and others},
  journal={Nature Communications (NC)},
  volume={16},
  number={1},
  pages={955},
  year={2025},
}

@article{xia2025discrete,
  title={Discrete Pneumatic-Tendon-Coupled Actuators with Interconnected Air Circuit for Untethered Soft Robots},
  author={Xia, Jiutian and Huang, Jie and Fu, Shiling and Qu, Jingting and Mo, Liyan and Li, Yunquan and Ren, Tao and Yang, Yang and Li, Yujia and Liu, Hao},
  journal={Advanced Intelligent Systems},
  volume={7},
  number={4},
  year={2025},
}

@book{birglen2008underactuated,
  title     = {Underactuated Robotic Hands},
  author    = {Birglen, Lionel and Lalibert{\'e}, Thierry and Gosselin, Cl{\'e}ment},
  year      = {2008},
  publisher = {Springer},
  address   = {Berlin, Heidelberg},
  series    = {Springer Tracts in Advanced Robotics},
  volume    = {40},
}

@article{palli2014dexmart,
  title={The DEXMART hand: Mechatronic design and experimental evaluation of synergy-based control for human-like grasping},
  author={Palli, Gianluca and Melchiorri, Claudio and Vassura, Gabriele and Scarcia, Umberto and Moriello, Lorenzo and Berselli, Giovanni and Cavallo, Alberto and De Maria, Giuseppe and Natale, Ciro and Pirozzi, Salvatore and others},
  journal=IJRR,
  volume={33},
  number={5},
  pages={799--824},
  year={2014},
}

@misc{shadowrobot2005,
  author       = {Shadow Robot Company},
  title        = {ShadowRobot Dexterous Hand},
  year         = {2005},
  howpublished = {\url{https://www.shadowrobot.com/dexterous-hand-series/}},
}

@inproceedings{bridgwater2012robonaut,
  title={The robonaut 2 hand-designed to do work with tools},
  author={Bridgwater, Lyndon B and Ihrke, CA and Diftler, Myron A and Abdallah, Muhammad E and Radford, Nicolaus A and Rogers, JM and Yayathi, S and Askew, R Scott and Linn, D Marty},
  booktitle=ICRA,
  pages={3425--3430},
  year={2012},
}

@article{grebenstein2012hand,
  title={The hand of the DLR hand arm system: Designed for interaction},
  author={Grebenstein, Markus and Chalon, Maxime and Friedl, Werner and Haddadin, Sami and Wimb{\"o}ck, Thomas and Hirzinger, Gerd and Siegwart, Roland},
  journal=IJRR,
  volume={31},
  number={13},
  pages={1531--1555},
  year={2012},
}

@inproceedings{tu2023posefusion,
  title={Posefusion: Robust object-in-hand pose estimation with selectlstm},
  author={Tu, Yuyang and Jiang, Junnan and Li, Shuang and Hendrich, Norman and Li, Miao and Zhang, Jianwei},
  booktitle=IROS,
  pages={6839--6846},
  year={2023},
}

@misc{schunk_svh,
  author       = {SCHUNK GmbH \& Co. KG},
  title        = {SVH 5-Finger Gripping Hand},
  howpublished = {\url{https://schunk.com/de_en/gripping-systems/highlights/svh/}},
  year         = {2025},
}

@inproceedings{ribeiro2023modeling,
  title={Modeling and realistic simulation of a dexterous robotic hand: SVH hand use-case},
  author={Ribeiro, Francisco M and Correia, Tiago and Lima, Jos{\'e} and Gon{\c{c}}alves, Gil and Pinto, V{\'\i}tor H},
  booktitle={IEEE International Conference on Autonomous Robot Systems and Competitions (ICARSC)},
  pages={132--138},
  year={2023},
}

@article{kim2021integrated,
  title={Integrated linkage-driven dexterous anthropomorphic robotic hand},
  author={Kim, Uikyum and Jung, Dawoon and Jeong, Heeyoen and Park, Jongwoo and Jung, Hyun-Mok and Cheong, Joono and Choi, Hyouk Ryeol and Do, Hyunmin and Park, Chanhun},
  journal={Nature Communications (NC)},
  volume={12},
  number={1},
  pages={7177},
  year={2021},
}

@article{birglen2004kinetostatic,
  title={Kinetostatic analysis of underactuated fingers},
  author={Birglen, Lionel and Gosselin, Cl{\'e}ment M},
  journal=TROA,
  volume={20},
  number={2},
  pages={211--221},
  year={2004},
}

@article{birglen2006grasp,
  title={Grasp-state plane analysis of two-phalanx underactuated fingers},
  author={Birglen, Lionel and Gosselin, Cl{\'e}ment M},
  journal={Mechanism and Machine Theory},
  volume={41},
  number={7},
  pages={807--822},
  year={2006},
}

@article{yoon2017underactuated,
  title={Underactuated finger mechanism using contractible slider-cranks and stackable four-bar linkages},
  author={Yoon, Dukchan and Choi, Youngjin},
  journal=TMECH,
  volume={22},
  number={5},
  pages={2046--2057},
  year={2017},
}

@article{wan2025rapid,
  title={RAPID Hand: A Robust, Affordable, Perception-Integrated, Dexterous Manipulation Platform for Generalist Robot Autonomy},
  author={Wan, Zhaoliang and Bi, Zetong and Zhou, Zida and Ren, Hao and Zeng, Yiming and Li, Yihan and Qi, Lu and Yang, Xu and Yang, Ming-Hsuan and Cheng, Hui},
  journal={arXiv preprint arXiv:2506.07490},
  year={2025}
}

@inproceedings{higashimori2005new,
  title={A new four-fingered robot hand with dual turning mechanism},
  author={Higashimori, Mitsuru and Jeong, Hieyong and Ishii, Idaku and Kaneko, Makoto and Namiki, Akio and Ishikawa, Masatoshi},
  booktitle=ICRA,
  pages={2679--2684},
  year={2005},
}

@inproceedings{quan2013planetary,
  title={A planetary gear based underactuated self-adaptive robotic finger},
  author={Quan, Qiquan and Wang, Qingchuan and Deng, Zongquan and Jiang, Shengyuan and Hou, Xuyan and Tang, Dewei},
  booktitle=ROBIO,
  pages={1586--1591},
  year={2013},
}

@inproceedings{tahara2012externally,
  title={Externally sensorless dynamic regrasping and manipulation by a triple-fingered robotic hand with torsional fingertip joints},
  author={Tahara, Kenji and Maruta, Keigo and Kawamura, Akihiro and Yamamoto, Motoji},
  booktitle=ICRA,
  pages={3252--3257},
  year={2012},
}

@inproceedings{yuan2020design,
  title={Design of a roller-based dexterous hand for object grasping and within-hand manipulation},
  author={Yuan, Shenli and Epps, Austin D and Nowak, Jerome B and Salisbury, J Kenneth},
  booktitle=ICRA,
  pages={8870--8876},
  year={2020},
}

@article{tuthill2018proprioception,
  title={Proprioception},
  author={Tuthill, John C and Azim, Eiman},
  journal={Current Biology},
  volume={28},
  number={5},
  pages={R194--R203},
  year={2018},
}

@article{morgan2021towards,
  title={Towards generalized manipulation learning through grasp mechanics-based features and self-supervision},
  author={Morgan, Andrew S and Bircher, Walter G and Dollar, Aaron M},
  journal=TRO,
  volume={37},
  number={5},
  pages={1553--1569},
  year={2021},
}

@article{shi2017dynamic,
  title={Dynamic in-hand sliding manipulation},
  author={Shi, Jian and Woodruff, J Zachary and Umbanhowar, Paul B and Lynch, Kevin M},
  journal=TRO,
  volume={33},
  number={4},
  pages={778--795},
  year={2017},
}

@article{li2019development,
  title={Development of a neural network-based control system for the DLR-HIT II robot hand using leap motion},
  author={Li, Chunxu and Fahmy, Ashraf and Sienz, Johann},
  journal={IEEE Access},
  volume={7},
  pages={136914--136923},
  year={2019},
}

@inproceedings{saegusa2010self,
  title={Self-body discovery based on visuomotor coherence},
  author={Saegusa, Ryo and Metta, Giorgio and Sandini, Giulio},
  booktitle={International Conference on Human System Interaction},
  pages={356--362},
  year={2010},
}

@inproceedings{liu2008multisensory,
  title={Multisensory five-finger dexterous hand: The DLR/HIT Hand II},
  author={Liu, Hong and Wu, Ke and Meusel, Peter and Seitz, Nikolaus and Hirzinger, Gerd and Jin, MH and Liu, YW and Fan, SW and Lan, T and Chen, ZP},
  booktitle=IROS,
  pages={3692--3697},
  year={2008},
}

@inproceedings{friedl2011fas,
  title={FAS A flexible antagonistic spring element for a high performance over},
  author={Friedl, Werner and Chalon, Maxime and Reinecke, Jens and Grebenstein, Markus and others},
  booktitle=IROS,
  pages={1366--1372},
  year={2011},
}

@article{wan2017recent,
  title={Recent progresses on flexible tactile sensors},
  author={Wan, Yongbiao and Wang, Yan and Guo, Chuan Fei},
  journal={Materials Today Physics},
  volume={1},
  pages={61--73},
  year={2017},
}

@article{zhao2016optoelectronically,
  title={Optoelectronically innervated soft prosthetic hand via stretchable optical waveguides},
  author={Zhao, Huichan and O’brien, Kevin and Li, Shuo and Shepherd, Robert F},
  journal={Science Robotics},
  volume={1},
  number={1},
  year={2016},
}

@article{mao2024multimodal,
  title={Multimodal tactile sensing fused with vision for dexterous robotic housekeeping},
  author={Mao, Qian and Liao, Zijian and Yuan, Jinfeng and Zhu, Rong},
  journal={Nature Communications (NC)},
  volume={15},
  number={1},
  pages={6871},
  year={2024},
}

@article{seong2025soft,
  title={Soft Sensors via Conductive Textile Stitching: Enabling Strain, Tactile, and Volumetric Sensing},
  author={Seong, Jihun and Lee, Ju-Hee and Han, Min-Woo},
  journal={Advanced Materials Technologies},
  volume={10},
  number={6},
  pages={2401306},
  year={2025},
}

@article{song2025fabric,
  title={A Fabric-Based Multimodal Flexible Tactile Sensor With Precise Sensing and Discrimination Capabilities for Pressure-Proximity-Magnetic Field Signals},
  author={Song, Mengya and Liu, Qiongzhen and Xu, Xiao and Wang, Bo and Lu, Ying and Yang, Liyan and Liu, Xue and Wang, Yuedan and Li, Mufang and Wang, Dong},
  journal={Advanced Functional Materials},
  volume={35},
  number={19},
  pages={2420445},
  year={2025},
}

@article{rostamian2022texture,
  title={Texture recognition based on multi-sensory integration of proprioceptive and tactile signals},
  author={Rostamian, Behnam and Koolani, MohammadReza and Abdollahzade, Pouya and Lankarany, Milad and Falotico, Egidio and Amiri, Mahmood and V. Thakor, Nitish},
  journal={Scientific Reports},
  volume={12},
  number={1},
  pages={21690},
  year={2022},
}

@inproceedings{ogahara2003wire,
  title={A wire-driven miniature five fingered robot hand using elastic elements as joints},
  author={Ogahara, Yoichi and Kawato, Yusuke and Takemura, Kenjiro and Maeno, Takashi},
  booktitle=IROS,
  volume={3},
  pages={2672--2677},
  year={2003},
}

@inproceedings{gilday2020vision,
  title={A Vision-Based Collocated Actuation-Sensing Scheme for a Compliant Tendon-Driven Robotic Hand.},
  author={Gilday, Kieran and Thuruthel, Thomas George and Iida, Fumiya},
  booktitle={RoboSoft},
  pages={760--765},
  year={2020}
}

@article{chen202518,
  title={An 18-DOF hand integrating force--position multimodal perception using a monocular camera},
  author={Chen, Shiwei and Li, Jiapeng and Deng, Zhiming and Wang, Peiji and Wei, Cheng and Cao, Xibin},
  journal={Nature Communications (NC)},
  volume={16},
  number={1},
  pages={6801},
  year={2025},
}

@article{wang2023neuromorphic,
  title={Neuromorphic sensorimotor loop embodied by monolithically integrated, low-voltage, soft e-skin},
  author={Wang, Weichen and Jiang, Yuanwen and Zhong, Donglai and Zhang, Zhitao and Choudhury, Snehashis and Lai, Jian-Cheng and Gong, Huaxin and Niu, Simiao and Yan, Xuzhou and Zheng, Yu and others},
  journal={Science},
  volume={380},
  number={6646},
  pages={735--742},
  year={2023},
}

@inproceedings{huang2019continuous,
  title={Continuous relaxation of symbolic planner for one-shot imitation learning},
  author={Huang, De-An and Xu, Danfei and Zhu, Yuke and Garg, Animesh and Savarese, Silvio and Fei-Fei, Li and Niebles, Juan Carlos},
  booktitle=IROS,
  pages={2635--2642},
  year={2019},
}

@article{bai2025towards,
  title={Towards a unified understanding of robot manipulation: A comprehensive survey},
  author={Bai, Shuanghao and Song, Wenxuan and Chen, Jiayi and Ji, Yuheng and Zhong, Zhide and Yang, Jin and Zhao, Han and Zhou, Wanqi and Zhao, Wei and Li, Zhe and others},
  journal={arXiv preprint arXiv:2510.10903},
  year={2025}
}

@inproceedings{song2023llm,
  title={Llm-planner: Few-shot grounded planning for embodied agents with large language models},
  author={Song, Chan Hee and Wu, Jiaman and Washington, Clayton and Sadler, Brian M and Chao, Wei-Lun and Su, Yu},
  booktitle=ICCV,
  pages={2998--3009},
  year={2023}
}

@article{liu2009modular,
  title={The modular multisensory DLR-HIT-Hand: hardware and software architecture},
  author={Liu, Hong and Meusel, Peter and Hirzinger, Gerd and Jin, Minghe and Liu, Yiwei and Xie, Zongwu},
  journal=TMECH,
  volume={13},
  number={4},
  pages={461--469},
  year={2009},
}

@article{yu2022dexterous,
  title={Dexterous manipulation for multi-fingered robotic hands with reinforcement learning: A review},
  author={Yu, Chunmiao and Wang, Peng},
  journal={Frontiers in Neurorobotics},
  volume={16},
  pages={861825},
  year={2022},
}

@article{fang2022tactonet,
  title={Tactonet: Tactile ordinal network based on unimodal probability for object hardness classification},
  author={Fang, Senlin and Yi, Zhengkun and Mi, Tingting and Zhou, Zhenning and Ye, Chaoxiang and Shang, Wanfeng and Xu, Tiantian and Wu, Xinyu},
  journal=TASE,
  volume={20},
  number={4},
  pages={2784--2794},
  year={2022},
}

@article{liu2024tactclnet,
  title={TactCLNet: Tactile continual learning network based on generative replay for object hardness recognition},
  author={Liu, Yiwen and Yi, Zhengkun and Fang, Senlin and Zhang, Yupo and Wan, Feng and Yang, Zhi-Xin and Lu, Xu and Wu, Xinyu},
  journal=TASE,
  year={2024},
}

@article{popov2017data,
  title={Data-efficient deep reinforcement learning for dexterous manipulation},
  author={Popov, Ivaylo and Heess, Nicolas and Lillicrap, Timothy and Hafner, Roland and Barth-Maron, Gabriel and Vecerik, Matej and Lampe, Thomas and Tassa, Yuval and Erez, Tom and Riedmiller, Martin},
  journal={arXiv preprint arXiv:1704.03073},
  year={2017}
}

@article{khandate2023sampling,
  title={Sampling-based exploration for reinforcement learning of dexterous manipulation},
  author={Khandate, Gagan and Shang, Siqi and Chang, Eric T and Saidi, Tristan Luca and Liu, Yang and Dennis, Seth Matthew and Adams, Johnson and Ciocarlie, Matei},
  journal={arXiv preprint arXiv:2303.03486},
  year={2023}
}

@inproceedings{gupta2021reset,
  title={Reset-free reinforcement learning via multi-task learning: Learning dexterous manipulation behaviors without human intervention},
  author={Gupta, Abhishek and Yu, Justin and Zhao, Tony Z and Kumar, Vikash and Rovinsky, Aaron and Xu, Kelvin and Devlin, Thomas and Levine, Sergey},
  booktitle=ICRA,
  pages={6664--6671},
  year={2021},
}

@article{xu2022dexterous,
  title={Dexterous manipulation from images: Autonomous real-world rl via substep guidance},
  author={Xu, Kelvin and Hu, Zheyuan and Doshi, Ria and Rovinsky, Aaron and Kumar, Vikash and Gupta, Abhishek and Levine, Sergey},
  journal={arXiv preprint arXiv:2212.09902},
  year={2022}
}

@inproceedings{qin2023dexpoint,
  title={Dexpoint: Generalizable point cloud reinforcement learning for sim-to-real dexterous manipulation},
  author={Qin, Yuzhe and Huang, Binghao and Yin, Zhao-Heng and Su, Hao and Wang, Xiaolong},
  booktitle=CORL,
  pages={594--605},
  year={2023},
}

@article{kannan2023deft,
  title={Deft: Dexterous fine-tuning for real-world hand policies},
  author={Kannan, Aditya and Shaw, Kenneth and Bahl, Shikhar and Mannam, Pragna and Pathak, Deepak},
  journal={arXiv preprint arXiv:2310.19797},
  year={2023}
}

@article{ze2023h,
  title={H-InDex: Visual reinforcement learning with hand-informed representations for dexterous manipulation},
  author={Ze, Yanjie and Liu, Yuyao and Shi, Ruizhe and Qin, Jiaxin and Yuan, Zhecheng and Wang, Jiashun and Xu, Huazhe},
  journal=NeurIPS,
  volume={36},
  pages={74394--74409},
  year={2023}
}

@inproceedings{yuan2024robot,
  title={Robot synesthesia: In-hand manipulation with visuotactile sensing},
  author={Yuan, Ying and Che, Haichuan and Qin, Yuzhe and Huang, Binghao and Yin, Zhao-Heng and Lee, Kang-Won and Wu, Yi and Lim, Soo-Chul and Wang, Xiaolong},
  booktitle=ICRA,
  pages={6558--6565},
  year={2024},
}

@article{chen2023bi,
  title={Bi-dexhands: Towards human-level bimanual dexterous manipulation},
  author={Chen, Yuanpei and Geng, Yiran and Zhong, Fangwei and Ji, Jiaming and Jiang, Jiechuang and Lu, Zongqing and Dong, Hao and Yang, Yaodong},
  journal=TPAMI,
  volume={46},
  number={5},
  pages={2804--2818},
  year={2023},
}

@article{zhou2025dydexhandover,
  title={DyDexHandover: Human-like Bimanual Dynamic Dexterous Handover using RGB-only Perception},
  author={Zhou, Haoran and You, Yangwei and Wang, Shuaijun},
  journal={arXiv preprint arXiv:2509.17350},
  year={2025}
}

@article{sivakumar2022robotic,
  title={Robotic telekinesis: Learning a robotic hand imitator by watching humans on youtube},
  author={Sivakumar, Aravind and Shaw, Kenneth and Pathak, Deepak},
  journal={arXiv preprint arXiv:2202.10448},
  year={2022}
}

@misc{wu2022ILAD,
    title={Learning Generalizable Dexterous Manipulation from Human Grasp Affordance},
    author={Wu, Yueh-Hua and Wang, Jiashun and Wang, Xiaolong},
    year={2022},
    archivePrefix={arXiv}
 }

@article{arunachalam2022dexterous,
  title={Dexterous imitation made easy: A learning-based framework for efficient dexterous manipulation},
  author={Arunachalam, Sridhar Pandian and Silwal, Sneha and Evans, Ben and Pinto, Lerrel},
  journal={arXiv preprint arXiv:2203.13251},
  year={2022}
}

@inproceedings{han2023utility,
  title={On the utility of koopman operator theory in learning dexterous manipulation skills},
  author={Han, Yunhai and Xie, Mandy and Zhao, Ye and Ravichandar, Harish},
  booktitle=CORL,
  pages={106--126},
  year={2023},
}

@ARTICLE{chen2025dexforce,
  author={Chen, Claire and Yu, Zhongchun and Choi, Hojung and Cutkosky, Mark and Bohg, Jeannette},
  journal=RA-L, 
  title={DexForce: Extracting Force-Informed Actions From Kinesthetic Demonstrations for Dexterous Manipulation}, 
  year={2025},
  volume={10},
  number={6},
  pages={6416-6423}
 }

@article{yu2025robotic,
  title={Robotic in-hand manipulation for large-range precise object movement: The rgmc champion solution},
  author={Yu, Mingrui and Jiang, Yongpeng and Chen, Chen and Jia, Yongyi and Li, Xiang},
  journal=RA-L,
  year={2025},
}

@inproceedings{guzey2025bridging,
  title={Bridging the human to robot dexterity gap through object-oriented rewards},
  author={Guzey, Irmak and Dai, Yinlong and Savva, Georgy and Bhirangi, Raunaq and Pinto, Lerrel},
  booktitle=ICRA,
  pages={3344--3351},
  year={2025},
}

@inproceedings{chen2025vividex,
  title={Vividex: Learning vision-based dexterous manipulation from human videos},
  author={Chen, Zerui and Chen, Shizhe and Arlaud, Etienne and Laptev, Ivan and Schmid, Cordelia},
  booktitle=ICRA,
  pages={3336--3343},
  year={2025}
}

@article{kim2024openvla,
  title={Openvla: An open-source vision-language-action model},
  author={Kim, Moo Jin and Pertsch, Karl and Karamcheti, Siddharth and Xiao, Ted and Balakrishna, Ashwin and Nair, Suraj and Rafailov, Rafael and Foster, Ethan and Lam, Grace and Sanketi, Pannag and others},
  journal={arXiv preprint arXiv:2406.09246},
  year={2024}
}

@article{team2024octo,
  title={Octo: An open-source generalist robot policy},
  author={Team, Octo Model and Ghosh, Dibya and Walke, Homer and Pertsch, Karl and Black, Kevin and Mees, Oier and Dasari, Sudeep and Hejna, Joey and Kreiman, Tobias and Xu, Charles and others},
  journal={arXiv preprint arXiv:2405.12213},
  year={2024}
}

@article{luo2025being,
  title={Being-h0: vision-language-action pretraining from large-scale human videos},
  author={Luo, Hao and Feng, Yicheng and Zhang, Wanpeng and Zheng, Sipeng and Wang, Ye and Yuan, Haoqi and Liu, Jiazheng and Xu, Chaoyi and Jin, Qin and Lu, Zongqing},
  journal={arXiv preprint arXiv:2507.15597},
  year={2025}
}

@article{zhen20243d,
  title={3d-vla: A 3d vision-language-action generative world model},
  author={Zhen, Haoyu and Qiu, Xiaowen and Chen, Peihao and Yang, Jincheng and Yan, Xin and Du, Yilun and Hong, Yining and Gan, Chuang},
  journal={arXiv preprint arXiv:2403.09631},
  year={2024}
}

@article{liu2025robodexvlm,
  title={Robodexvlm: Visual language model-enabled task planning and motion control for dexterous robot manipulation},
  author={Liu, Haichao and Guo, Sikai and Mai, Pengfei and Cao, Jiahang and Li, Haoang and Ma, Jun},
  journal={arXiv preprint arXiv:2503.01616},
  year={2025}
}

@article{chen2025villa,
  title={Villa-x: enhancing latent action modeling in vision-language-action models},
  author={Chen, Xiaoyu and Wei, Hangxing and Zhang, Pushi and Zhang, Chuheng and Wang, Kaixin and Guo, Yanjiang and Yang, Rushuai and Wang, Yucen and Xiao, Xinquan and Zhao, Li and others},
  journal={arXiv preprint arXiv:2507.23682},
  year={2025}
}

@article{wen2025dexvla,
  title={Dexvla: Vision-language model with plug-in diffusion expert for general robot control},
  author={Wen, Junjie and Zhu, Yichen and Li, Jinming and Tang, Zhibin and Shen, Chaomin and Feng, Feifei},
  journal={arXiv preprint arXiv:2502.05855},
  year={2025}
}

@article{cheng2025omnivtla,
  title={OmniVTLA: Vision-Tactile-Language-Action Model with Semantic-Aligned Tactile Sensing},
  author={Cheng, Zhengxue and Zhang, Yiqian and Zhang, Wenkang and Li, Haoyu and Wang, Keyu and Song, Li and Zhang, Hengdi},
  journal={arXiv preprint arXiv:2508.08706},
  year={2025}
}

@article{wagenmaker2025steering,
  title={Steering Your Diffusion Policy with Latent Space Reinforcement Learning},
  author={Wagenmaker, Andrew and Nakamoto, Mitsuhiko and Zhang, Yunchu and Park, Seohong and Yagoub, Waleed and Nagabandi, Anusha and Gupta, Abhishek and Levine, Sergey},
  journal={arXiv preprint arXiv:2506.15799},
  year={2025}
}

@inproceedings{wu2025tacdiffusion,
  title={Tacdiffusion: Force-domain diffusion policy for precise tactile manipulation},
  author={Wu, Yansong and Chen, Zongxie and Wu, Fan and Chen, Lingyun and Zhang, Liding and Bing, Zhenshan and Swikir, Abdalla and Haddadin, Sami and Knoll, Alois},
  booktitle=ICRA,
  pages={11831--11837},
  year={2025},
}

@inproceedings{liang2025dexhanddiff,
  title={Dexhanddiff: Interaction-aware diffusion planning for adaptive dexterous manipulation},
  author={Liang, Zhixuan and Mu, Yao and Wang, Yixiao and Chen, Tianxing and Shao, Wenqi and Zhan, Wei and Tomizuka, Masayoshi and Luo, Ping and Ding, Mingyu},
  booktitle={Proceedings of the Computer Vision and Pattern Recognition Conference},
  pages={1745--1755},
  year={2025}
}

@article{zhang2025manidext,
  title={Manidext: Hand-object manipulation synthesis via continuous correspondence embeddings and residual-guided diffusion},
  author={Zhang, Jiajun and Zhang, Yuxiang and An, Liang and Li, Mengcheng and Zhang, Hongwen and Hu, Zonghai and Liu, Yebin},
  journal=TPAMI,
  year={2025},
}

@article{guo2024prediction,
  title={Prediction with action: Visual policy learning via joint denoising process},
  author={Guo, Yanjiang and Hu, Yucheng and Zhang, Jianke and Wang, Yen-Jen and Chen, Xiaoyu and Lu, Chaochao and Chen, Jianyu},
  journal=NeurIPS,
  volume={37},
  pages={112386--112410},
  year={2024}
}

@inproceedings{chen2021trajectotree,
  title={Trajectotree: Trajectory optimization meets tree search for planning multi-contact dexterous manipulation},
  author={Chen, Claire and Culbertson, Preston and Lepert, Marion and Schwager, Mac and Bohg, Jeannette},
  booktitle=IROS,
  pages={8262--8268},
  year={2021},
}

@article{luo2024robust,
  title={Robust tube-based MPC with smooth computation for dexterous robot manipulation},
  author={Luo, Yu and Ji, Tianying and Sun, Fuchun and Sima, Qie and Liu, Huaping and Jing, Mingxuan and Zhang, Jianwei},
  journal={Science China Information Sciences},
  volume={67},
  number={11},
  pages={212207},
  year={2024},
  publisher={Springer}
}

@article{xu2024letac,
  title={LeTac-MPC: Learning model predictive control for tactile-reactive grasping},
  author={Xu, Zhengtong and She, Yu},
  journal=TRO,
  year={2024},
}

@article{lum2024dextrah,
  title={Dextrah-g: Pixels-to-action dexterous arm-hand grasping with geometric fabrics},
  author={Lum, Tyler Ga Wei and Matak, Martin and Makoviychuk, Viktor and Handa, Ankur and Allshire, Arthur and Hermans, Tucker and Ratliff, Nathan D and Van Wyk, Karl},
  journal={arXiv preprint arXiv:2407.02274},
  year={2024}
}

@inproceedings{caggiano2023myodex,
  title={Myodex: a generalizable prior for dexterous manipulation},
  author={Caggiano, Vittorio and Dasari, Sudeep and Kumar, Vikash},
  booktitle=ICML,
  pages={3327--3346},
  year={2023},
}

@inproceedings{bicchi2000robotic,
  title={Robotic grasping and contact: A review},
  author={Bicchi, Antonio and Kumar, Vijay},
  booktitle=ICRA,
  volume={1},
  pages={348--353},
  year={2000},
}

@inproceedings{huber2024domain,
  title={Domain randomization for sim2real transfer of automatically generated grasping datasets},
  author={Huber, Johann and H{\'e}l{\'e}non, Fran{\c{c}}ois and Watrelot, Hippolyte and Amar, Fa{\"\i}z Ben and Doncieux, St{\'e}phane},
  booktitle=ICRA,
  pages={4112--4118},
  year={2024},
}

@article{liu2024realdex,
  title={Realdex: Towards human-like grasping for robotic dexterous hand},
  author={Liu, Yumeng and Yang, Yaxun and Wang, Youzhuo and Wu, Xiaofei and Wang, Jiamin and Yao, Yichen and Schwertfeger, S{\"o}ren and Yang, Sibei and Wang, Wenping and Yu, Jingyi and others},
  journal={arXiv preprint arXiv:2402.13853},
  year={2024}
}

@article{song2025overview,
  title={An overview of learning-based dexterous grasping: recent advances and future directions},
  author={Song, Xu and Li, Yongyao and Zhang, Yunfan and Liu, Yufei and Jiang, Lei},
  journal={Artificial Intelligence Review},
  volume={58},
  number={10},
  pages={1--44},
  year={2025},
}

@article{li2024comprehensive,
  title={A comprehensive review of robot intelligent grasping based on tactile perception},
  author={Li, Tong and Yan, Yuhang and Yu, Chengshun and An, Jing and Wang, Yifan and Chen, Gang},
  journal=RCIM,
  volume={90},
  pages={102792},
  year={2024},
}

@article{billard2019trends,
  title={Trends and challenges in robot manipulation},
  author={Billard, Aude and Kragic, Danica},
  journal={Science},
  volume={364},
  number={6446},
  pages={eaat8414},
  year={2019},
}

@article{wang2023dexterous,
  title={Dexterous robotic manipulation using deep reinforcement learning and knowledge transfer for complex sparse reward-based tasks},
  author={Wang, Qiang and Sanchez, Francisco Roldan and McCarthy, Robert and Bulens, David Cordova and McGuinness, Kevin and O'Connor, Noel and W{\"u}thrich, Manuel and Widmaier, Felix and Bauer, Stefan and Redmond, Stephen J},
  journal={Expert Systems},
  volume={40},
  number={6},
  pages={e13205},
  year={2023},
}

@article{yuan2024cross,
  title={Cross-embodiment dexterous grasping with reinforcement learning},
  author={Yuan, Haoqi and Zhou, Bohan and Fu, Yuhui and Lu, Zongqing},
  journal={arXiv preprint arXiv:2410.02479},
  year={2024}
}

@inproceedings{luo2024serl,
  title={Serl: A software suite for sample-efficient robotic reinforcement learning},
  author={Luo, Jianlan and Hu, Zheyuan and Xu, Charles and Tan, You Liang and Berg, Jacob and Sharma, Archit and Schaal, Stefan and Finn, Chelsea and Gupta, Abhishek and Levine, Sergey},
  booktitle=ICRA,
  pages={16961--16969},
  year={2024},
}

@inproceedings{schoettler2020deep,
  title={Deep reinforcement learning for industrial insertion tasks with visual inputs and natural rewards},
  author={Schoettler, Gerrit and Nair, Ashvin and Luo, Jianlan and Bahl, Shikhar and Ojea, Juan Aparicio and Solowjow, Eugen and Levine, Sergey},
  booktitle=IROS,
  pages={5548--5555},
  year={2020},
}

@article{li2024grasp,
  title={Grasp with push policy for multi-finger dexterity hand based on deep reinforcement learning},
  author={Li, Baojiang and Qiu, Shengjie and Bai, Jibo and Wang, Haiyan and Wang, Bin and Zhang, Zhekai and Li, Liang and Wang, Xichao},
  journal={Applied Soft Computing},
  volume={167},
  pages={112365},
  year={2024},
}

@article{hu2023grasping,
  title={Grasping living objects with adversarial behaviors using inverse reinforcement learning},
  author={Hu, Zhe and Zheng, Yu and Pan, Jia},
  journal=TRO,
  volume={39},
  number={2},
  pages={1151--1163},
  year={2023},
}

@article{qin2022one,
  title={From one hand to multiple hands: Imitation learning for dexterous manipulation from single-camera teleoperation},
  author={Qin, Yuzhe and Su, Hao and Wang, Xiaolong},
  journal=RA-L,
  volume={7},
  number={4},
  pages={10873--10881},
  year={2022},
}

@inproceedings{qin2022dexmv,
  title={Dexmv: Imitation learning for dexterous manipulation from human videos},
  author={Qin, Yuzhe and Wu, Yueh-Hua and Liu, Shaowei and Jiang, Hanwen and Yang, Ruihan and Fu, Yang and Wang, Xiaolong},
  booktitle=ECCV,
  pages={570--587},
  year={2022},
}

@article{liu2025immimic,
  title={Immimic: Cross-domain imitation from human videos via mapping and interpolation},
  author={Liu, Yangcen and Shin, Woo Chul and Han, Yunhai and Chen, Zhenyang and Ravichandar, Harish and Xu, Danfei},
  journal={arXiv preprint arXiv:2509.10952},
  year={2025}
}

@article{li2025softgrasp,
  title={SoftGrasp: Adaptive grasping for dexterous hand based on multimodal imitation learning},
  author={Li, Yihong and Guo, Ce and Ren, Junkai and Chen, Bailiang and Cheng, Chuang and Zhang, Hui and Lu, Huimin},
  journal={Biomimetic Intelligence and Robotics},
  volume={5},
  number={2},
  pages={100217},
  year={2025},
}

@article{lin2025pp,
  title={PP-Tac: Paper Picking Using Tactile Feedback in Dexterous Robotic Hands},
  author={Lin, Pei and Huang, Yuzhe and Li, Wanlin and Ma, Jianpeng and Xiao, Chenxi and Jiao, Ziyuan},
  journal={arXiv preprint arXiv:2504.16649},
  year={2025}
}

@article{wang2024neural,
  title={Neural attention field: Emerging point relevance in 3d scenes for one-shot dexterous grasping},
  author={Wang, Qianxu and Deng, Congyue and Lum, Tyler Ga Wei and Chen, Yuanpei and Yang, Yaodong and Bohg, Jeannette and Zhu, Yixin and Guibas, Leonidas},
  journal={arXiv preprint arXiv:2410.23039},
  year={2024}
}

@article{li2025single,
  title={A single-demonstration guided manipulation learning with dexterous hand},
  author={Li, Jianwen and Lv, Yinglan and Lin, Xiangbo and Hang, Jinglue and Li, Xuanheng and Sun, Yi},
  journal=EAAI,
  volume={159},
  pages={111606},
  year={2025},
}

@article{wei2023generalized,
  title={Generalized anthropomorphic functional grasping with minimal demonstrations},
  author={Wei, Wei and Wang, Peng and Wang, Sizhe},
  journal={arXiv preprint arXiv:2303.17808},
  year={2023}
}

@inproceedings{wang2024cyberdemo,
  title={Cyberdemo: Augmenting simulated human demonstration for real-world dexterous manipulation},
  author={Wang, Jun and Qin, Yuzhe and Kuang, Kaiming and Korkmaz, Yigit and Gurumoorthy, Akhilan and Su, Hao and Wang, Xiaolong},
  booktitle=CVPR,
  pages={17952--17963},
  year={2024}
}

@article{zhong2025dexgraspvla,
  title={Dexgraspvla: A vision-language-action framework towards general dexterous grasping},
  author={Zhong, Yifan and Huang, Xuchuan and Li, Ruochong and Zhang, Ceyao and Chen, Zhang and Guan, Tianrui and Zeng, Fanlian and Lui, Ka Num and Ye, Yuyao and Liang, Yitao and others},
  journal={arXiv preprint arXiv:2502.20900},
  year={2025}
}

@article{deng2025graspvla,
  title={Graspvla: a grasping foundation model pre-trained on billion-scale synthetic action data},
  author={Deng, Shengliang and Yan, Mi and Wei, Songlin and Ma, Haixin and Yang, Yuxin and Chen, Jiayi and Zhang, Zhiqi and Yang, Taoyu and Zhang, Xuheng and Zhang, Wenhao and others},
  journal={arXiv preprint arXiv:2505.03233},
  year={2025}
}

@article{pan2024vision,
  title={Vision-language-action model and diffusion policy switching enables dexterous control of an anthropomorphic hand},
  author={Pan, Cheng and Junge, Kai and Hughes, Josie},
  journal={arXiv preprint arXiv:2410.14022},
  year={2024}
}

@inproceedings{wu2023learning,
  title={Learning generalizable dexterous manipulation from human grasp affordance},
  author={Wu, Yueh-Hua and Wang, Jiashun and Wang, Xiaolong},
  booktitle=CORL,
  pages={618--629},
  year={2023},
}

@article{weng2024dexdiffuser,
  title={Dexdiffuser: Generating dexterous grasps with diffusion models},
  author={Weng, Zehang and Lu, Haofei and Kragic, Danica and Lundell, Jens},
  journal=RA-L,
  year={2024},
}

@inproceedings{zhao2024graingrasp,
  title={Graingrasp: Dexterous grasp generation with fine-grained contact guidance},
  author={Zhao, Fuqiang and Tsetserukou, Dzmitry and Liu, Qian},
  booktitle=ICRA,
  pages={6470--6476},
  year={2024},
}

@inproceedings{wang2024gendp,
  title={Gendp: 3d semantic fields for category-level generalizable diffusion policy},
  author={Wang, Yixuan and Yin, Guang and Huang, Binghao and Kelestemur, Tarik and Wang, Jiuguang and Li, Yunzhu},
  booktitle=CORL,
  volume={2},
  year={2024}
}

@article{chen2024springgrasp,
  title={Springgrasp: Synthesizing compliant, dexterous grasps under shape uncertainty},
  author={Chen, Sirui and Bohg, Jeannette and Liu, C Karen},
  journal={arXiv preprint arXiv:2404.13532},
  year={2024}
}

@article{makarova2025diffusionrl,
  title={DiffusionRL: Efficient Training of Diffusion Policies for Robotic Grasping Using RL-Adapted Large-Scale Datasets},
  author={Makarova, Maria and Liu, Qian and Tsetserukou, Dzmitry},
  journal={arXiv preprint arXiv:2505.18876},
  year={2025}
}

@article{wei2024d,
  title={D (r, o) grasp: A unified representation of robot and object interaction for cross-embodiment dexterous grasping},
  author={Wei, Zhenyu and Xu, Zhixuan and Guo, Jingxiang and Hou, Yiwen and Gao, Chongkai and Cai, Zhehao and Luo, Jiayu and Shao, Lin},
  journal={arXiv preprint arXiv:2410.01702},
  year={2024}
}

@inproceedings{xu2023unidexgrasp,
  title={Unidexgrasp: Universal robotic dexterous grasping via learning diverse proposal generation and goal-conditioned policy},
  author={Xu, Yinzhen and Wan, Weikang and Zhang, Jialiang and Liu, Haoran and Shan, Zikang and Shen, Hao and Wang, Ruicheng and Geng, Haoran and Weng, Yijia and Chen, Jiayi and others},
  booktitle=CVPR,
  pages={4737--4746},
  year={2023}
}

@inproceedings{wan2023unidexgrasp++,
  title={Unidexgrasp++: Improving dexterous grasping policy learning via geometry-aware curriculum and iterative generalist-specialist learning},
  author={Wan, Weikang and Geng, Haoran and Liu, Yun and Shan, Zikang and Yang, Yaodong and Yi, Li and Wang, He},
  booktitle=ICCV,
  pages={3891--3902},
  year={2023}
}

@article{wang2025d3grasp,
  title={D3Grasp: Diverse and Deformable Dexterous Grasping for General Objects},
  author={Wang, Keyu and Lu, Bingcong and Cheng, Zhengxue and Zhang, Hengdi and Song, Li},
  journal={arXiv preprint arXiv:2509.19892},
  year={2025}
}

@inproceedings{christen2022d,
  title={D-grasp: Physically plausible dynamic grasp synthesis for hand-object interactions},
  author={Christen, Sammy and Kocabas, Muhammed and Aksan, Emre and Hwangbo, Jemin and Song, Jie and Hilliges, Otmar},
  booktitle=CVPR,
  pages={20577--20586},
  year={2022}
}

@inproceedings{wang2025unigrasptransformer,
  title={Unigrasptransformer: Simplified policy distillation for scalable dexterous robotic grasping},
  author={Wang, Wenbo and Wei, Fangyun and Zhou, Lei and Chen, Xi and Luo, Lin and Yi, Xiaohan and Zhang, Yizhong and Liang, Yaobo and Xu, Chang and Lu, Yan and others},
  booktitle=CVPR,
  pages={12199--12208},
  year={2025}
}

@article{xiao2025designing,
  title={Designing Pin-pression Gripper and Learning its Dexterous Grasping with Online In-hand Adjustment},
  author={Xiao, Hewen and Liu, Xiuping and Zhao, Hang and Liu, Jian and Xu, Kai},
  journal=TOG,
  volume={44},
  number={4},
  pages={1--17},
  year={2025},
}

@article{singh2024dextrah,
  title={Dextrah-rgb: Visuomotor policies to grasp anything with dexterous hands},
  author={Singh, Ritvik and Allshire, Arthur and Handa, Ankur and Ratliff, Nathan and Van Wyk, Karl},
  journal={arXiv preprint arXiv:2412.01791},
  year={2024}
}

@inproceedings{zhou2023learning,
  title={Learning to grasp the ungraspable with emergent extrinsic dexterity},
  author={Zhou, Wenxuan and Held, David},
  booktitle=CORL,
  pages={150--160},
  year={2023},
  organization={PMLR}
}

@inproceedings{lin2025learning,
  title={Learning visuotactile skills with two multifingered hands},
  author={Lin, Toru and Zhang, Yu and Li, Qiyang and Qi, Haozhi and Yi, Brent and Levine, Sergey and Malik, Jitendra},
  booktitle=ICRA,
  pages={5637--5643},
  year={2025},
}

@article{chen2022towards,
  title={Towards human-level bimanual dexterous manipulation with reinforcement learning},
  author={Chen, Yuanpei and Wu, Tianhao and Wang, Shengjie and Feng, Xidong and Jiang, Jiechuan and Lu, Zongqing and McAleer, Stephen and Dong, Hao and Zhu, Song-Chun and Yang, Yaodong},
  journal=NeurIPS,
  volume={35},
  pages={5150--5163},
  year={2022}
}

@article{franzese2023interactive,
  title={Interactive imitation learning of bimanual movement primitives},
  author={Franzese, Giovanni and de Souza Rosa, Leandro and Verburg, Tim and Peternel, Luka and Kober, Jens},
  journal=TMECH,
  year={2023},
}

@inproceedings{li2025maniptrans,
  title={Maniptrans: Efficient dexterous bimanual manipulation transfer via residual learning},
  author={Li, Kailin and Li, Puhao and Liu, Tengyu and Li, Yuyang and Huang, Siyuan},
  booktitle=CVPR,
  pages={6991--7003},
  year={2025}
}

@inproceedings{liu2024taco,
  title={Taco: Benchmarking generalizable bimanual tool-action-object understanding},
  author={Liu, Yun and Yang, Haolin and Si, Xu and Liu, Ling and Li, Zipeng and Zhang, Yuxiang and Liu, Yebin and Yi, Li},
  booktitle=CVPR,
  pages={21740--21751},
  year={2024}
}

@article{zhou2024learning,
  title={Learning diverse bimanual dexterous manipulation skills from human demonstrations},
  author={Zhou, Bohan and Yuan, Haoqi and Fu, Yuhui and Lu, Zongqing},
  journal={arXiv preprint arXiv:2410.02477},
  year={2024}
}

@article{chi2024universal,
  title={Universal manipulation interface: In-the-wild robot teaching without in-the-wild robots},
  author={Chi, Cheng and Xu, Zhenjia and Pan, Chuer and Cousineau, Eric and Burchfiel, Benjamin and Feng, Siyuan and Tedrake, Russ and Song, Shuran},
  journal={arXiv preprint arXiv:2402.10329},
  year={2024}
}

@inproceedings{jiang2025dexmimicgen,
  title={Dexmimicgen: Automated data generation for bimanual dexterous manipulation via imitation learning},
  author={Jiang, Zhenyu and Xie, Yuqi and Lin, Kevin and Xu, Zhenjia and Wan, Weikang and Mandlekar, Ajay and Fan, Linxi Jim and Zhu, Yuke},
  booktitle=ICRA,
  pages={16923--16930},
  year={2025},
}

@inproceedings{gbagbe2024bi,
  title={Bi-vla: Vision-language-action model-based system for bimanual robotic dexterous manipulations},
  author={Gbagbe, Koffivi Fid{\`e}le and Cabrera, Miguel Altamirano and Alabbas, Ali and Alyunes, Oussama and Lykov, Artem and Tsetserukou, Dzmitry},
  booktitle={IEEE International Conference on Systems, Man, and Cybernetics (SMC)},
  pages={2864--2869},
  year={2024},
}

@article{liu2024rdt,
  title={Rdt-1b: a diffusion foundation model for bimanual manipulation},
  author={Liu, Songming and Wu, Lingxuan and Li, Bangguo and Tan, Hengkai and Chen, Huayu and Wang, Zhengyi and Xu, Ke and Su, Hang and Zhu, Jun},
  journal={arXiv preprint arXiv:2410.07864},
  year={2024}
}

@inproceedings{khan2025shake,
  title={Shake-vla: Vision-language-action model-based system for bimanual robotic manipulations and liquid mixing},
  author={Khan, Muhamamd Haris and Asfaw, Selamawit and Iarchuk, Dmitrii and Cabrera, Miguel Altamirano and Moreno, Luis and Tokmurziyev, Issatay and Tsetserukou, Dzmitry},
  booktitle={ACM/IEEE International Conference on Human-Robot Interaction (HRI)},
  pages={1393--1397},
  year={2025},
}

@inproceedings{ma2024hierarchical,
  title={Hierarchical diffusion policy for kinematics-aware multi-task robotic manipulation},
  author={Ma, Xiao and Patidar, Sumit and Haughton, Iain and James, Stephen},
  booktitle=CVPR,
  pages={18081--18090},
  year={2024}
}

@article{huang20253dvitaclearningfinegrainedmanipulation,
  title={3D-ViTac: Learning Fine-Grained Manipulation with Visuo-Tactile Sensing}, 
  author={Binghao Huang and Yixuan Wang and Xinyi Yang and Yiyue Luo and Yunzhu Li},
  year={2024},
  journal={arXiv preprint arXiv:2410.24091}
}

@inproceedings{fan2023arctic,
  title={ARCTIC: A dataset for dexterous bimanual hand-object manipulation},
  author={Fan, Zicong and Taheri, Omid and Tzionas, Dimitrios and Kocabas, Muhammed and Kaufmann, Manuel and Black, Michael J and Hilliges, Otmar},
  booktitle=CVPR,
  pages={12943--12954},
  year={2023}
}

@article{wu2024robomind,
  title={Robomind: Benchmark on multi-embodiment intelligence normative data for robot manipulation},
  author={Wu, Kun and Hou, Chengkai and Liu, Jiaming and Che, Zhengping and Ju, Xiaozhu and Yang, Zhuqin and Li, Meng and Zhao, Yinuo and Xu, Zhiyuan and Yang, Guang and others},
  journal={arXiv preprint arXiv:2412.13877},
  year={2024}
}

@inproceedings{christen2023learning,
  title={Learning human-to-robot handovers from point clouds},
  author={Christen, Sammy and Yang, Wei and P{\'e}rez-D’Arpino, Claudia and Hilliges, Otmar and Fox, Dieter and Chao, Yu-Wei},
  booktitle=CVPR,
  pages={9654--9664},
  year={2023}
}

@inproceedings{bahl2023affordances,
  title={Affordances from human videos as a versatile representation for robotics},
  author={Bahl, Shikhar and Mendonca, Russell and Chen, Lili and Jain, Unnat and Pathak, Deepak},
  booktitle=CVPR,
  pages={13778--13790},
  year={2023}
}

@article{jawaid2025openego,
  title={Openego: A large-scale multimodal egocentric dataset for dexterous manipulation},
  author={Jawaid, Ahad and Xiang, Yu},
  journal={arXiv preprint arXiv:2509.05513},
  year={2025}
}

@article{luo2025precise,
  title={Precise and dexterous robotic manipulation via human-in-the-loop reinforcement learning},
  author={Luo, Jianlan and Xu, Charles and Wu, Jeffrey and Levine, Sergey},
  journal={Science Robotics},
  volume={10},
  number={105},
  year={2025},
}

@article{lin2025sim,
  title={Sim-to-real reinforcement learning for vision-based dexterous manipulation on humanoids},
  author={Lin, Toru and Sachdev, Kartik and Fan, Linxi and Malik, Jitendra and Zhu, Yuke},
  journal={arXiv preprint arXiv:2502.20396},
  year={2025}
}

@article{tang2025mimicfunc,
  title={Mimicfunc: Imitating tool manipulation from a single human video via functional correspondence},
  author={Tang, Chao and Xiao, Anxing and Deng, Yuhong and Hu, Tianrun and Dong, Wenlong and Zhang, Hanbo and Hsu, David and Zhang, Hong},
  journal={arXiv preprint arXiv:2508.13534},
  year={2025}
}

@inproceedings{nagabandi2020deep,
  title={Deep dynamics models for learning dexterous manipulation},
  author={Nagabandi, Anusha and Konolige, Kurt and Levine, Sergey and Kumar, Vikash},
  booktitle=CORL,
  pages={1101--1112},
  year={2020},
  organization={PMLR}
}

@article{xu2025hierarchical,
  title={Hierarchical reinforcement learning for articulated tool manipulation with multifingered hand},
  author={Xu, Wei and Zhao, Yanchao and Guo, Weichao and Sheng, Xinjun},
  journal={arXiv preprint arXiv:2507.06822},
  year={2025}
}

@article{atar2025hand,
  title={In-Hand Manipulation of Articulated Tools with Dexterous Robot Hands with Sim-to-Real Transfer},
  author={Atar, Soofiyan and Huang, Daniel and Richter, Florian and Yip, Michael},
  journal={arXiv preprint arXiv:2509.23075},
  year={2025}
}

@article{liu2023learning,
  title={Learning to design and use tools for robotic manipulation},
  author={Liu, Ziang and Tian, Stephen and Guo, Michelle and Liu, C Karen and Wu, Jiajun},
  journal={arXiv preprint arXiv:2311.00754},
  year={2023}
}

@inproceedings{zitkovich2023rt,
  title={Rt-2: Vision-language-action models transfer web knowledge to robotic control},
  author={Zitkovich, Brianna and Yu, Tianhe and Xu, Sichun and Xu, Peng and Xiao, Ted and Xia, Fei and Wu, Jialin and Wohlhart, Paul and Welker, Stefan and Wahid, Ayzaan and others},
  booktitle=CORL,
  pages={2165--2183},
  year={2023},
}

@article{black2024pi0visionlanguageactionflowmodel,
  title={$\pi_0$: A Vision-Language-Action Flow Model for General Robot Control}, 
  author={Kevin Black and Noah Brown and Danny Driess and Adnan Esmail and Michael Equi and Chelsea Finn and Niccolo Fusai and Lachy Groom and Karol Hausman and Brian Ichter and Szymon Jakubczak and Tim Jones and Liyiming Ke and Sergey Levine and Adrian Li-Bell and Mohith Mothukuri and Suraj Nair and Karl Pertsch and Lucy Xiaoyang Shi and James Tanner and Quan Vuong and Anna Walling and Haohuan Wang and Ury Zhilinsky},
  journal={arXiv preprint arXiv:2410.24164},
  year={2023}
}

@article{yin2025dexteritygen,
  title={Dexteritygen: Foundation controller for unprecedented dexterity},
  author={Yin, Zhao-Heng and Wang, Changhao and Pineda, Luis and Hogan, Francois and Bodduluri, Krishna and Sharma, Akash and Lancaster, Patrick and Prasad, Ishita and Kalakrishnan, Mrinal and Malik, Jitendra and others},
  journal=RSS,
  year={2025}
}

@article{fang2020learning,
  title={Learning task-oriented grasping for tool manipulation from simulated self-supervision},
  author={Fang, Kuan and Zhu, Yuke and Garg, Animesh and Kurenkov, Andrey and Mehta, Viraj and Fei-Fei, Li and Savarese, Silvio},
  journal=IJRR,
  volume={39},
  number={2-3},
  pages={202--216},
  year={2020},
}

@INPROCEEDINGS{chen2025bodex,
  author={Chen, Jiayi and Ke, Yubin and Wang, He},
  booktitle=ICRA, 
  title={BODex: Scalable and Efficient Robotic Dexterous Grasp Synthesis Using Bilevel Optimization}, 
  year={2025},
  volume={},
  number={},
  pages={01-08}
  }

@article{wu2025cedex,
  title={CEDex: Cross-Embodiment Dexterous Grasp Generation at Scale from Human-like Contact Representations}, 
  author={Zhiyuan Wu and Rolandos Alexandros Potamias and Xuyang Zhang and Zhongqun Zhang and Jiankang Deng and Shan Luo},
  journal={arXiv preprint arXiv:2509.24661},
  year={2025}
}

@inproceedings{ye2025dex1b,
  title={Dex1B: Learning with 1B Demonstrations for Dexterous Manipulation},
  author={Ye, Jianglong and Wang, Keyi and Yuan, Chengjing and Yang, Ruihan and Li, Yiquan and Zhu, Jiyue and Qin, Yuzhe and Zou, Xueyan and Wang, Xiaolong},
  booktitle=RSS,
  year={2025}
}

@article{wang2024dexcap,
  title={Dexcap: Scalable and portable mocap data collection system for dexterous manipulation},
  author={Wang, Chen and Shi, Haochen and Wang, Weizhuo and Zhang, Ruohan and Fei-Fei, Li and Liu, C Karen},
  journal={arXiv preprint arXiv:2403.07788},
  year={2024}
}

@inproceedings{hang2024dexfuncgrasp,
  title={DexFuncGrasp: A Robotic Dexterous Functional Grasp Dataset Constructed from a Cost-Effective Real-Simulation Annotation System.},
  author={Hang, Jinglue and Lin, Xiangbo and Zhu, Tianqiang and Li, Xuanheng and Wu, Rina and Ma, Xiaohong and Sun, Yi},
  booktitle=AAAI,
  pages={10306-10313},
  year={2024}
}

@article{zhong2025dexgrasp,
        title={DexGrasp Anything: Towards Universal Robotic Dexterous Grasping with Physics Awareness},
        author={Zhong, Yiming and Jiang, Qi and Yu, Jingyi and Ma, Yuexin},
        journal={arXiv preprint arXiv:2503.08257},
        year={2025}
      }

@article{wang2022dexgraspnet,
  title={Dexgraspnet: A large-scale robotic dexterous grasp dataset for general objects based on simulation},
  author={Wang, Ruicheng and Zhang, Jialiang and Chen, Jiayi and Xu, Yinzhen and Li, Puhao and Liu, Tengyu and Wang, He},
  journal={arXiv preprint arXiv:2210.02697},
  year={2022}
}

@article{Zhang2025dextog,
   title={DexTOG: Learning Task-Oriented Dexterous Grasp With Language Condition},
   volume={10},
   ISSN={2377-3774},
   number={2},
   journal=RA-L,
   author={Zhang, Jieyi and Xu, Wenqiang and Yu, Zhenjun and Xie, Pengfei and Tang, Tutian and Lu, Cewu},
   year={2025},
}

@misc{li2022gendexgrasp,
    title        = {GenDexGrasp: Generalizable Dexterous Grasping},
    author       = {Li, Puhao and Liu, Tengyu and Li, Yuyang and Zhu, Yixin and Yang, Yaodong and Huang, Siyuan},
    howpublished = {url{https://arxiv.org/abs/2210.00722}},
    year         = {2022}
}

@inproceedings{liu2025vtdexmanip,
    title={VTDexManip: A Dataset and Benchmark for Visual-tactile Pretraining and Dexterous Manipulation with Reinforcement Learning},
    author={Qingtao Liu and Yu Cui and Zhengnan Sun and Gaofeng Li and Jiming Chen and Qi Ye},
    booktitle=ICLR,
    year={2025},
}

@inproceedings{garcia2018first,
  title={First-person hand action benchmark with rgb-d videos and 3d hand pose annotations},
  author={Garcia-Hernando, Guillermo and Yuan, Shanxin and Baek, Seungryul and Kim, Tae-Kyun},
  booktitle=CVPR,
  pages={409--419},
  year={2018}
}

@inproceedings{brahmbhatt2020contactpose,
  title={ContactPose: A dataset of grasps with object contact and hand pose},
  author={Brahmbhatt, Samarth and Tang, Chengcheng and Twigg, Christopher D and Kemp, Charles C and Hays, James},
  booktitle=ECCV,
  pages={361--378},
  year={2020},
}

@inproceedings{chao2021dexycb,
  title={DexYCB: A benchmark for capturing hand grasping of objects},
  author={Chao, Yu-Wei and Yang, Wei and Xiang, Yu and Molchanov, Pavlo and Handa, Ankur and Tremblay, Jonathan and Narang, Yashraj S and Van Wyk, Karl and Iqbal, Umar and Birchfield, Stan and others},
  booktitle=CVPR,
  pages={9044--9053},
  year={2021}
}

@inproceedings{yang2022oakink,
  title={Oakink: A large-scale knowledge repository for understanding hand-object interaction},
  author={Yang, Lixin and Li, Kailin and Zhan, Xinyu and Wu, Fei and Xu, Anran and Liu, Liu and Lu, Cewu},
  booktitle=CVPR,
  pages={20953--20962},
  year={2022}
}

@inproceedings{todorov2012mujoco,
  title={Mujoco: A physics engine for model-based control},
  author={Todorov, Emanuel and Erez, Tom and Tassa, Yuval},
  booktitle=IROS,
  pages={5026--5033},
  year={2012},
}

@inproceedings{fang2024rh20t,
  title={Rh20t: A comprehensive robotic dataset for learning diverse skills in one-shot},
  author={Fang, Hao-Shu and Fang, Hongjie and Tang, Zhenyu and Liu, Jirong and Wang, Chenxi and Wang, Junbo and Zhu, Haoyi and Lu, Cewu},
  booktitle=ICRA,
  pages={653--660},
  year={2024},
}

@inproceedings{o2024open,
  title={Open x-embodiment: Robotic learning datasets and rt-x models: Open x-embodiment collaboration 0},
  author={O’Neill, Abby and Rehman, Abdul and Maddukuri, Abhiram and Gupta, Abhishek and Padalkar, Abhishek and Lee, Abraham and Pooley, Acorn and Gupta, Agrim and Mandlekar, Ajay and Jain, Ajinkya and others},
  booktitle=ICRA,
  pages={6892--6903},
  year={2024},
}

@article{vemprala2024chatgpt,
  title={Chatgpt for robotics: Design principles and model abilities},
  author={Vemprala, Sai H and Bonatti, Rogerio and Bucker, Arthur and Kapoor, Ashish},
  journal={IEEE Access},
  volume={12},
  pages={55682--55696},
  year={2024},
}

@article{brohan2022rt,
  title={Rt-1: Robotics transformer for real-world control at scale},
  author={Brohan, Anthony and Brown, Noah and Carbajal, Justice and Chebotar, Yevgen and Dabis, Joseph and Finn, Chelsea and Gopalakrishnan, Keerthana and Hausman, Karol and Herzog, Alex and Hsu, Jasmine and others},
  journal={arXiv preprint arXiv:2212.06817},
  year={2022}
}

@article{andrychowicz2020learning,
  title={Learning dexterous in-hand manipulation},
  author={Andrychowicz, OpenAI: Marcin and Baker, Bowen and Chociej, Maciek and Jozefowicz, Rafal and McGrew, Bob and Pachocki, Jakub and Petron, Arthur and Plappert, Matthias and Powell, Glenn and Ray, Alex and others},
  journal=IJRR,
  volume={39},
  number={1},
  pages={3--20},
  year={2020},
}

@article{zhao2025high,
  title={A high-fidelity simulation framework for grasping stability analysis in human casualty manipulation},
  author={Zhao, Qianwen and Roy, Rajarshi and Spurlock, Chad and Lister, Kevin and Wang, Long},
  journal=TMRB,
  year={2025},
}

@article{rothemund2021hasel,
  title={HASEL artificial muscles for a new generation of lifelike robots—recent progress and future opportunities},
  author={Rothemund, Philipp and Kellaris, Nicholas and Mitchell, Shane K and Acome, Eric and Keplinger, Christoph},
  journal={Advanced Materials},
  volume={33},
  number={19},
  pages={2003375},
  year={2021},
}

@inproceedings{bao2023dexart,
  title={Dexart: Benchmarking generalizable dexterous manipulation with articulated objects},
  author={Bao, Chen and Xu, Helin and Qin, Yuzhe and Wang, Xiaolong},
  booktitle=CVPR,
  pages={21190--21200},
  year={2023}
}

@article{Fang2024MOKAOR,
  title={MOKA: Open-World Robotic Manipulation through Mark-Based Visual Prompting},
  author={Kuan Fang and Fangchen Liu and Pieter Abbeel and Sergey Levine},
  journal=RSS,
  year={2024}
}

@article{Li2025HAMSTERHA,
  title={HAMSTER: Hierarchical Action Models For Open-World Robot Manipulation},
  author={Yi Li and Yuquan Deng and Jesse Zhang and Joel Jang and Marius Memmel and Caelan Reed Garrett and Fabio Ramos and Dieter Fox and AnqiLi and AbhishekGupta and Ankit Goyal and Nvidia},
  journal=ICLR,
  year={2025}
}

@article{Intelligence202505AV,
  title={$pi0.5$: a Vision-Language-Action Model with Open-World Generalization},
  author={Physical Intelligence and Kevin Black and Noah Brown and James Darpinian and Karan Dhabalia and Danny Driess and Adnan Esmail and Michael Equi and Chelsea Finn and Niccolo Fusai and Manuel Y. Galliker and Dibya Ghosh and Lachy Groom and Karol Hausman and Brian Ichter and Szymon Jakubczak and Tim Jones and Liyiming Ke and Devin LeBlanc and Sergey Levine and Adrian Li-Bell and Mohith Mothukuri and Suraj Nair and Karl Pertsch and Allen Z. Ren and Lucy Xiaoyang Shi and Laura Smith and Jost Tobias Springenberg and Kyle Stachowicz and James Tanner and Quan Vuong and Homer Rich Walke and Anna Walling and Haohuan Wang and Lili Yu and Ury Zhilinsky},
  journal=CoRL,
  year={2025}
}

@inproceedings{bai2025retrdex,
    author    = {Fengshuo Bai and Yu Li and Jie Chu and Tawei Chou and Runchuan Zhu and Ying Wen and Yaodong Yang and Yuanpei Chen},
    title     = {Retrieval Dexterity: Efficient Object Retrieval in Clutters with Dexterous Hand},
    booktitle = {arXiv preprint arXiv:2502.18423},
    year      = {2025}
}

@article{zhao2026tele,
  title={Tele-Catch: Adaptive Teleoperation for Dexterous Dynamic 3D Object Catching},
  author={Zhao, Weiguang and Dong, Junting and Zhang, Rui and Li, Kailin and Zhao, Qin and Huang, Kaizhu},
  journal={arXiv preprint arXiv:2603.28427},
  year={2026}
}

@article{zhao20243d,
  title={3D-CDRGP: Towards Cross-Device Robotic Grasping Policy in 3D Open World},
  author={Zhao, Weiguang and Jiang, Chenru and Zhang, Chengrui and Sun, Jie and Yan, Yuyao and Zhang, Rui and Huang, Kaizhu},
  journal={arXiv preprint arXiv:2411.18133},
  year={2024}
}

@inproceedings{zhao2023divide,
  title={Divide and conquer: 3d point cloud instance segmentation with point-wise binarization},
  author={Zhao, Weiguang and Yan, Yuyao and Yang, Chaolong and Ye, Jianan and Yang, Xi and Huang, Kaizhu},
  booktitle=ICCV,
  pages={562--571},
  year={2023}
}

@inproceedings{zhao2025bfanet,
  title={Bfanet: Revisiting 3d semantic segmentation with boundary feature analysis},
  author={Zhao, Weiguang and Zhang, Rui and Wang, Qiufeng and Cheng, Guangliang and Huang, Kaizhu},
  booktitle=CVPR,
  pages={29395--29405},
  year={2025}
}

@article{zhang2024survey,
  title={A survey of trustworthy federated learning: Issues, solutions, and challenges},
  author={Zhang, Yifei and Zeng, Dun and Luo, Jinglong and Fu, Xinyu and Chen, Guanzhong and Xu, Zenglin and King, Irwin},
  journal={ACM Transactions on Intelligent Systems and Technology (TIST)},
  volume={15},
  number={6},
  pages={1--47},
  year={2024},
}

@inproceedings{zhang2023survey,
  title={A survey of trustworthy federated learning with perspectives on security, robustness and privacy},
  author={Zhang, Yifei and Zeng, Dun and Luo, Jinglong and Xu, Zenglin and King, Irwin},
  booktitle={ACM web conference (WWW)},
  pages={1167--1176},
  year={2023}
}

@article{yin2025disturbance,
  title={Disturbance compensation control for humanoid robot hand driven by tendon-sheath based on disturbance observer},
  author={Yin, Meng and Wang, Haozhe and Shang, Dongyang and Li, Ming and Xu, Tiantian and Wu, Xinyu},
  journal=TASE,
  year={2025},
}

@article{hao2025learn,
  title={Learn-gen-plan: Bridging the gap between vision language models and real-world long-horizon dexterous manipulations},
  author={Hao, Peng and Cui, Shaowei and Wei, Junhang and Lu, Tao and Cai, Yinghao and Wang, Shuo},
  journal=TASE,
  year={2025},
}

@article{li2025language,
  title={Language-guided dexterous functional grasping by llm generated grasp functionality and synergy for humanoid manipulation},
  author={Li, Zhuo and Liu, Junjia and Li, Zhihao and Dong, Zhipeng and Teng, Tao and Ou, Yongsheng and Caldwell, Darwin and Chen, Fei},
  journal=TASE,
  volume={22},
  pages={10506--10519},
  year={2025},
}

\end{document}